%% file: main.tex
\documentclass[a4paper,fleqn]{cas-dc}

\usepackage[authoryear]{natbib}
\usepackage[english]{babel}
\usepackage[autostyle=true]{csquotes} 
\usepackage{graphicx}
\usepackage{indentfirst}
\PassOptionsToPackage{table,xcdraw}{xcolor}
\usepackage{multirow, multicol}
\usepackage{tabularx}
\usepackage{geometry}
\usepackage{rotating}
\usepackage{longtable}
\usepackage{lscape}
\usepackage{ragged2e}
\usepackage{makeidx}
\usepackage{caption}
\usepackage{float}
\usepackage{booktabs}
\usepackage{makecell}
\usepackage{array}
\usepackage{amsmath}
\usepackage{subcaption}
\usepackage{algorithm}
\usepackage{algorithmic}
\usepackage{xcolor}
\usepackage{soul}

\def\tsc#1{\csdef{#1}{\textsc{\lowercase{#1}}\xspace}}
\tsc{WGM}
\tsc{QE}
\tsc{EP}
\tsc{PMS}
\tsc{BEC}
\tsc{DE}

\begin{document}
\let\WriteBookmarks\relax
\def\floatpagepagefraction{1}
\def\textpagefraction{.001}

\shorttitle{FedMSE: Semi-supervised federated learning approach for IoT network intrusion detection}

\shortauthors{Nguyen et~al.}

\title [mode = title]{FedMSE: Semi-supervised federated learning approach for IoT network intrusion detection}                      



%
\author[1,2]{Van Tuan Nguyen}[type=editor]

\cormark[1]


\ead{vantuan.nguyen@lqdtu.edu.vn}


\credit{Conceptualization; Methodology; Formal analysis; Validation; Writing - Original draft}

\affiliation[1]{organization={Japan Advanced Institute of Science and Technology},
    addressline={1-1, Asahidai}, 
    city={Nomi},
    postcode={923-1292}, 
    state={Ishikawa},
    country={Japan}}

\affiliation[2]{organization={Le Quy Don Technical University},
    addressline={236, Hoang Quoc Viet, Co Nhue 1},
    postcode={10000},
    state={Hanoi},
    country={Vietnam}
}
\author[1]{Razvan Beuran}

\credit{Supervision; Writing - Review \& editing}


\cortext[cor1]{Corresponding author}



\begin{abstract}
This paper proposes a novel federated learning approach for improving IoT network intrusion detection.
The rise of IoT has expanded the cyber attack surface, making traditional centralized machine learning methods insufficient due to concerns about data availability, computational resources, transfer costs, and especially privacy preservation. 
A semi-supervised federated learning model was developed to overcome these issues, combining the Shrink Autoencoder and Centroid one-class classifier (SAE-CEN). This approach enhances the performance of intrusion detection by effectively representing normal network data and accurately identifying anomalies in the decentralized strategy. Additionally, a mean square error-based aggregation algorithm (MSEAvg) was introduced to improve global model performance by prioritizing more accurate local models. The results obtained in our experimental setup, which uses various settings relying on the N-BaIoT dataset and Dirichlet distribution, demonstrate significant improvements in real-world heterogeneous IoT networks in detection accuracy from 93.98$\pm$2.90 to 97.30$\pm$0.49, reduced learning costs when requiring only 50\% of gateways participating in the training process, and robustness in large-scale networks.

\end{abstract}




\maketitle
\input{Sections/1-Introduction}
\input{Sections/2-Background}

\input{Sections/3-RelatedWork}
\input{Sections/4-Methodology}

\input{Sections/5-0-Evaluation}
\input{Sections/6-Conclusion}



\printcredits

\bibliographystyle{cas-model2-names}


\input{main.bbl}




\end{document}

%% file: Sections/1-Introduction.tex
\section{Introduction}
The Internet of Things (IoT) has emerged as a trendy technology due to its enormous potential in a variety of industries, including transportation \citep{iot-transportation}, healthcare \citep{iot-healcare}, and smart cities \citep{iot-smartcity}. IoT is the interconnected network of numerous physical objects, often known as things \citep{iot-definition}. IoT systems are increasing dramatically with new, sophisticated, and modern devices being produced every day. The data generated by an enormous number of interconnected devices is heterogeneous, diverse, and complex. This increases the cyber attack surface affected by several novel IoT anomalies \citep{iot-attack}, such as botnet, DoS, DDoS, and Spoofing. Thus, IoT intrusion detection is a very essential task in new-fashioned IoT networks. 

To secure system networks and privacy in deployment against cyber risks, an intrusion detection system (IDS) by analyzing and monitoring network traffic is one of the most efficient solutions \citep{iot-definition, Fed-ANIDS}, these approaches are also known as anomaly detection systems in some research. 
Modern IDSs that leverage machine learning (ML) techniques to detect unseen threats in IoT networks have been studied and adopted widely \citep{iot-definition,iot-attack,fl-iot}. 
They can be categorized into three main types: supervised learning, semi-supervised learning, and unsupervised learning \citep{anomaly-detection-survey}. Supervised learning works with the assumption that all available data is labeled for building the model. Any new data are fed into the trained model and classified as normal or abnormal. 
The semi-supervised learning methods suppose that only normal data is available to construct the model. They infer the unseen data based on the deviation from normal data.
The last type is unsupervised learning, which does not require labeled training data and works under the implicit assumption that the normal data points are far from anomalous ones. Semi-supervised and unsupervised learning techniques are often used more widely in intrusion detection due to their ability to detect unknown anomalies.
In addition to developing effective data learning models, it is also very important to choose and define learning and deployment strategies to face new challenges when IoT networks continuously scale up. 
Traditional ML methods work in a centralized paradigm, in which one model is trained on a single server using all data collected from connected agents~\citep{fl-review}. In the modern IoT context, this approach depicts some challenges such as computational resource requirements for training and serving models; data latency when the data transmission distance may be hundreds, even thousands of miles from the server; data transfer cost; particularly data privacy preservation issues due to data leakage when sharing data among agents and server~\citep{fl-survey}, especially in network intrusion detection.
Network data packets contain extremely sensitive data information, such as device addresses, connection sessions, personal user information, system vulnerabilities, etc. The continuous exchange and centralized storage of this information in computer networks increases the risk of data leakage, leading to serious information incidents.
Therefore, Federated Learning (FL) has emerged recently as a cutting-edge approach to solving these problems. Rather than collecting data, training, and serving ML models in a single high-performance computer, FL takes advantage of the computing capabilities of the network client in the distributed mechanism to train the local model itself and protect the local data from being accessed directly from outside \citep{fl-survey}. 
Then, the local models will be aggregated to a global model in the server for serving by some methodologies. 

In recent years, many approaches have been released to take advantage of the power of FL to solve network anomaly detection \citep{diot,Fed-ANIDS,gru-ensemble,fed-feature-selection}. However, several issues have been making this task challenging, such as 
the traffic collected from large-scale IoT networks with different application scenarios is unbalanced and does not have sufficient abnormal data for the training process; 
IoT network is scalable, the kind of IoT devices is very diverse, and generated data is heterogeneous;
the aggregation method of the global model is still not optimal for getting a robust model to detect novel attacks effectively and perform efficiently with resource-constrained IoT devices. 

To tackle these limitations, in this article, we aim to propose FedMSE, a collaborative IoT network intrusion detection approach leveraging the Autoencoder-based feature representation method, Mean Squared Error, and FL. The source code was published on Github \footnote{https://github.com/dino-chiio/fedmse} together with the dataset prepared for this paper. FedMSE enables the strength of identifying unknown cyber threats of semi-supervised learning techniques and the secure learning process of FL. Each IoT gateway independently trains an intrusion detection model on its locally processed network data without sharing the data with a central server. We use a hybrid approach, constructed by Shrink Autoencoder (SAE) as a feature representation method and Centroid (CEN) as a one-class classifier \citep{sae-model}, to improve the performance of intrusion detection in local gateways. Then, we develop a novel federated aggregation algorithm based on Mean Squared Error to enhance the accuracy, efficiency, and robustness of decentralized learning. This new method considers the contribution of local models in the training process and prioritizes more accurate models. The experimental evaluation demonstrates the enhancement in IoT network intrusion detection of our proposed approach.

The major contributions of this article are the following:
\begin{enumerate}
    \item Enabling secure machine learning in IoT network intrusion detection using federated, and semi-supervised learning.
    \item Proposing a novel FL algorithm to enhance the ability to detect IoT anomalies in heterogeneous environments. 
    \item Performing extensive experiments using a famous IoT dataset to evaluate our proposed approach. The experimental results present an advancement in distributed IoT network intrusion detection compared to some popular methods. 
    \item Conducting a comprehensive analysis of the algorithm's robustness in both resource-constrained and large-scale IoT networks. This analysis highlights the practical applications of the proposed approach.
\end{enumerate}

The rest of this paper is organized as follows. The next section briefly introduces some fundamental background related to this research. Section \ref{sec:related-work} highlights some recent research in federated intrusion detection. Our proposed approach is described in Section \ref{sec:proposed-approach}. Section \ref{sec:evaluation} gives the settings of the experiment and analyzes the results given by the proposed approach. Finally, we conclude the paper and discuss future work in Section \ref{sec:conclusion}.

%% file: Sections/2-Background.tex
\section{Background}
\label{sec:background}
This section briefly introduces the background knowledge related to the intrusion detection system with federated learning.

\subsection{Federated learning}

\begin{figure*}[p]
  \centering
  \includegraphics[width=\linewidth]{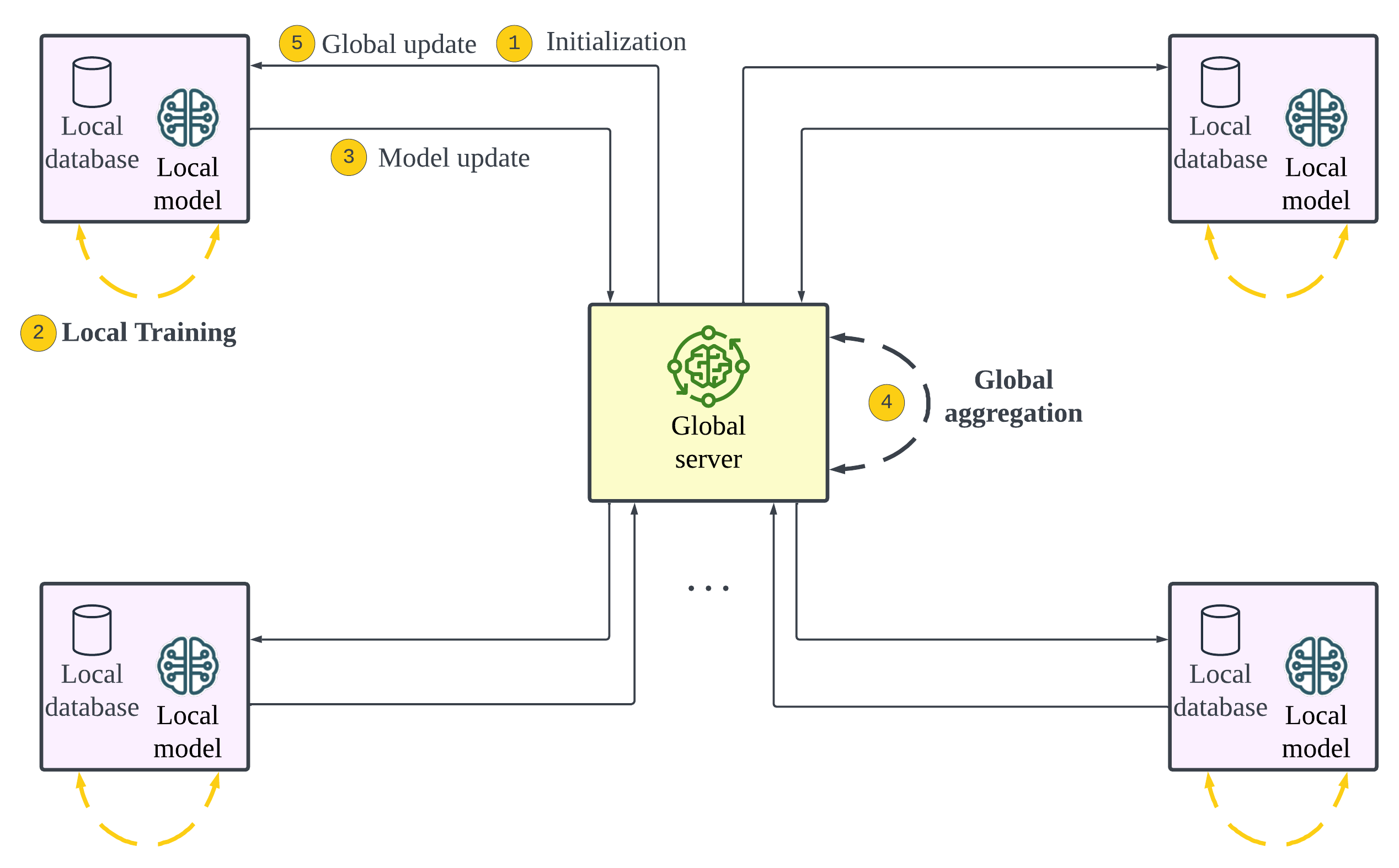} 
  \caption{Illustration of the Federated Learning mechanism.}
  \label{fig:fl-overview}
\end{figure*}

Federated Learning \citep{FL-proposal} is a novel approach to train machine learning models in a decentralized manner, aiming to address growing concerns about data privacy and the increasing computational power available on edge devices such as smartphones and IoT devices. FL enables the development of machine learning models by leveraging data distributed across multiple devices, referred to as clients or gateways in this research, without transferring the data to a central server. The model on each client device is called the local model, while the aggregated model, obtained by combining updates from clients, is known as the global model.

The training process is conducted in several consecutive communication rounds until the global model reaches the desired accuracy or meets the specified training criteria~\citep{fl-survey}. Figure $\ref{fig:fl-overview}$ depicts the steps of how a communication round occurs:

\begin{enumerate}
    \item \textit{Model initialization}: Firstly, a subset of gateways is selected in the global model with a selection ratio. A global model is initialized and sent to all selected gateways. This is the starting point for the training process.

    \item \textit{Local training}: Each gateway receives the global model and trains it using its local data. This results in a set of updated local models, each adapted to the data from its respective devices.

    \item \textit{Model update}: Once all selected devices have done the local training process, optimized weights are sent to the global server.

    \item \textit{Global model aggregation}: The server aggregates the updates from all local models to form a new global model that has the knowledge of all participants.

    \item \textit{Global update}: The updated global model is then sent back to all clients in the network for further prediction. This updated model now reflects a more comprehensive understanding based on the combined data from all gateways.

\end{enumerate}

The federated aggregation algorithm plays a crucial role in achieving the objectives of FL. Researchers have studied and proposed numerous algorithms for various applications. Among these, FedAvg \citep{FL-proposal} and FedProx~\citep{FedProx} are the most widely adopted and popular.

FedAvg is a straightforward algorithm that constructs a global model by calculating the weighted average of parameters from the client's models based on the number of data samples they hold. The client holding more data gets a higher weight in the aggregation process. The process occurs until the desired model is obtained. This algorithm is simple to implement and suitable for large-scale networks with a lot of participants \citep{Fed-ANIDS}. 

FedProx is similar to FedAvg but introduces a regularization term to each client's loss function, penalizing significant deviations of the client model parameters from the previous global model.

\begin{figure}[t]
  \centering
  \begin{subfigure}[t]{0.49\textwidth}
      \centering
      \includegraphics[width=\linewidth]{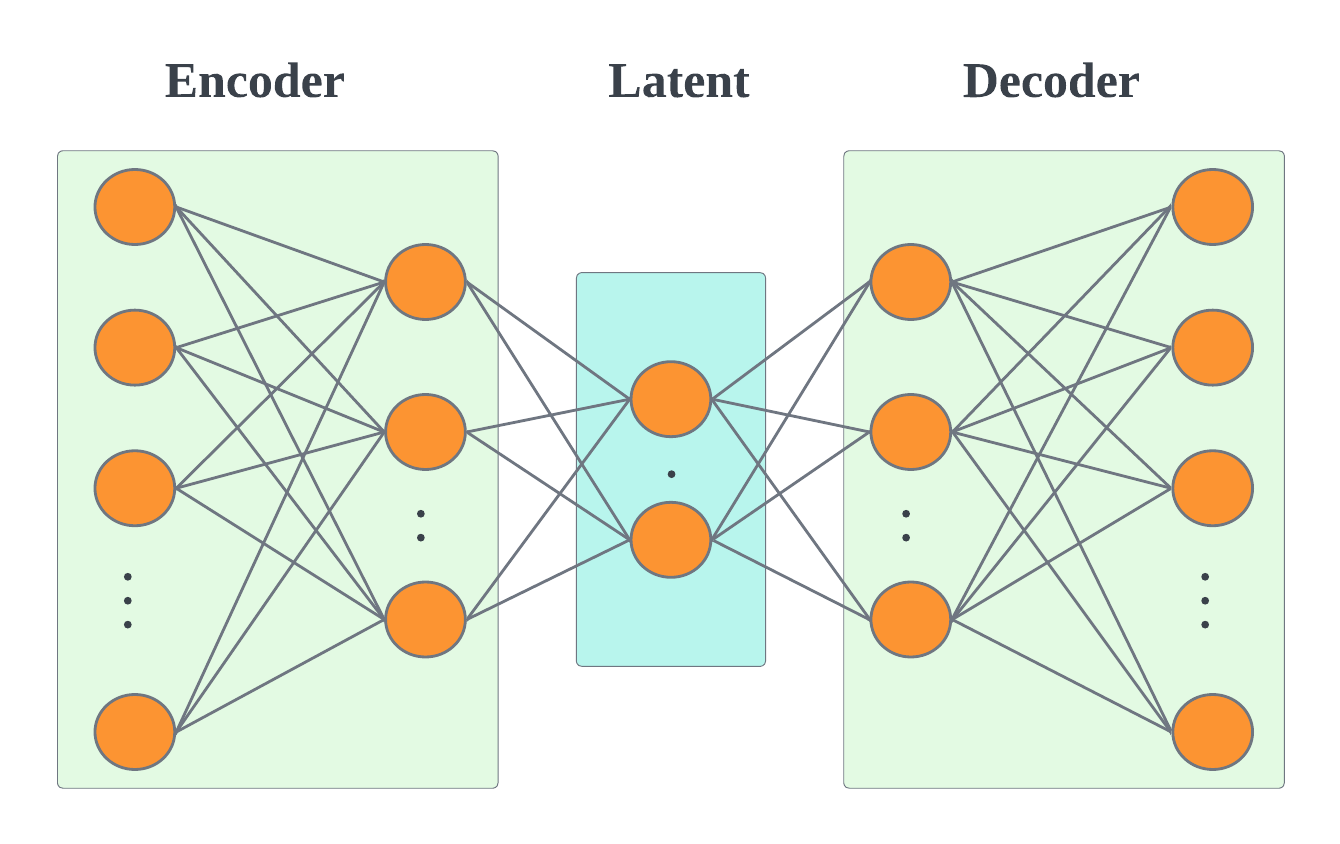}
      \captionsetup{justification=centering}
      \caption{}
      \label{fig:background-ae}
  \end{subfigure}

  \begin{subfigure}[t]{0.49\textwidth}
      \centering
      \includegraphics[width=\linewidth]{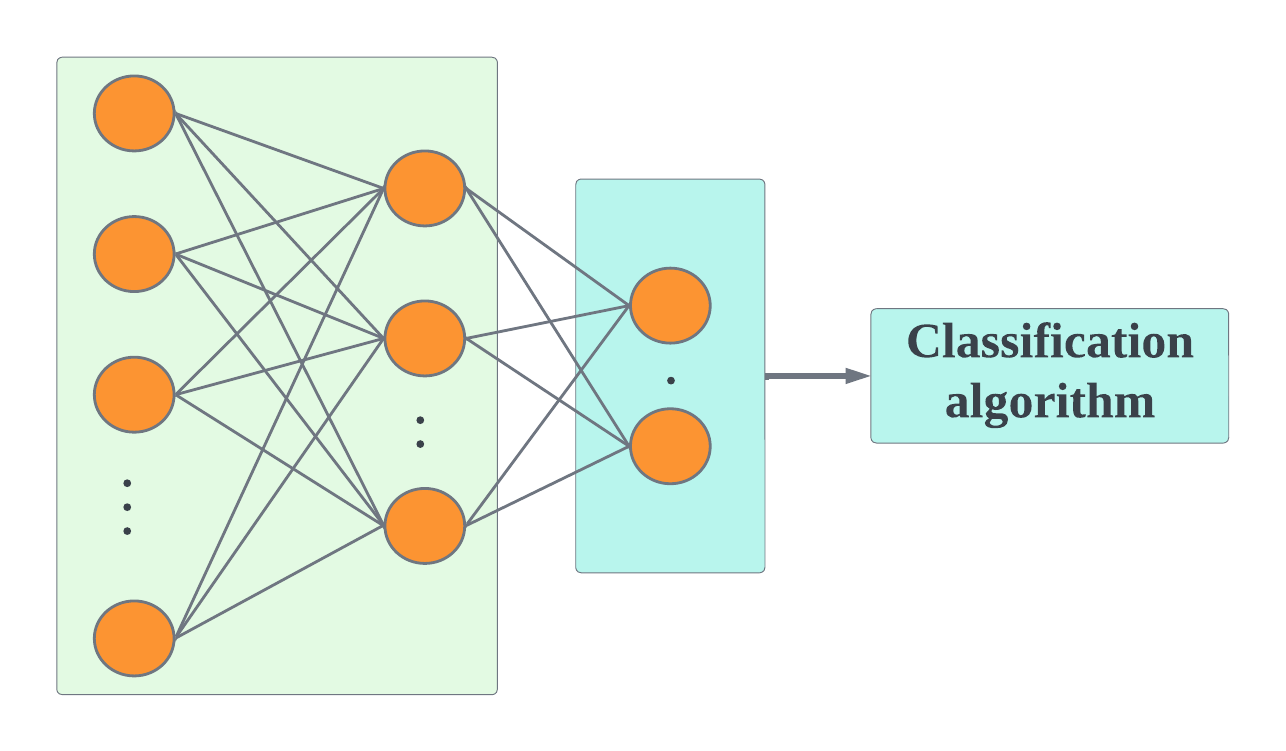}
      \captionsetup{justification=centering}
      \caption{}
      \label{fig:background-ae-rep}
  \end{subfigure}
  \caption{Architecture of (a) Autoencoder and (b) use of Autoencoder as feature representation for anomaly detection}
  \label{fig:combined-ae}
\end{figure}

\subsection{Autoencoder}
Autoencoder (AE) is a feed-forward neural network that consists of an encoder and a decoder (as shown in Figure \ref{fig:combined-ae}), trying to copy its input to its output \citep{deeplearningbook}. Let $X = \{x^{1}, \ldots, x^{n}\} \subset \mathbb{R}^n$ be the dataset. Let $\phi = (W, b)$, $\Phi = (W^{'}, b^{'})$ be weight matrices and bias vectors of the encoder and decoder, correspondingly; $e_\phi$ and $d_\Phi$ stand for encoder and decoder. The encoder compresses input data $x^{i}$ to latent representation $h^{i}$ in a new feature space of the latent layer, then the decoder maps it to the input feature space, getting reconstruction output $\hat{x}^{i}$. The AE can be formed as follows:

\begin{subequations}
\begin{align}
    h^i &= e_\phi \left( x^i \right) = act_e \left( W x^i + b \right) \label{eq:encoder} \\
    \hat{x}^i &= d_\Phi \left( h^i \right) = act_d \left( W' h^i + b' \right) \label{eq:decoder}
\end{align}
\end{subequations}
where $act_e$, and $act_d$ are activation functions of the encoder and decoder, respectively.

AEs are trained by minimizing the differences between the input data and the reconstruction data following Equation~\ref{eq:ae-loss} using the backpropagation technique:

\begin{equation}
\label{eq:ae-loss}
\mathcal{L}_{\text{AE}}(x^i, \hat{x}^i) = \frac{1}{n} \sum_{i=1} \| x^i - \hat{x}^i \|^2
\end{equation}
where $n$ is the number of data samples.

AEs are particularly useful for intrusion detection due to their ability to capture the underlying structure of normal data during training. Once trained on normal data, the AE will produce low reconstruction errors (Equation $\ref{eq:ae-loss}$) for inputs similar to the training data. Nevertheless, the AE finds it difficult to correctly recreate data that is anomalous, that is, data that differs greatly from the patterns discovered during training. Reconstruction errors can be used as a metric to classify anomalies. Higher errors indicate higher probabilities for anomalous.

In many applications, especially in network intrusion detection, AEs can be used for feature dimension reduction by compressing the original data into a lower-dimension space. The most important characteristics of original complex data can be extracted and represented more effectively in the latent layer, not only making other machine learning models learn data patterns more easily but also reducing the computational complexity in detection.

%% file: Sections/3-RelatedWork.tex
\section{Related work}
\label{sec:related-work}
In this section, we review some recent studies that motivate our research on federated learning for IoT network intrusion detection. 

As mentioned before, existing network intrusion detection approaches are facing some challenges. 
The scarcity of labeled anomalous data is one of the major issues in advanced intrusion detection. 
Despite the vast amount of network traffic data generated by modern and complex IoT networks, there is a notable lack of labeled abnormal data due to privacy issues, limitations of expert knowledge and labeling expenses~\citep{sae-model}. 
Anomalous network traffic includes computer communication related to system weaknesses and vulnerabilities. Thus, administrators tend to keep the data private to protect their network or client's sensitive information.
Moreover, IoT networks generate a huge amount of network data and intrusion behaviors become more sophisticated over time, leading to being more challenging to collect, process, and label anomalous traffic in real-world networks that require a significant amount of time and expert knowledge.
This absence makes it difficult to use supervised machine learning models to learn the characteristics of available intrusive data. 
To address this, some semi-supervised learning techniques, such as OCSVM~\citep{ocsvm}, and LOF \citep{lof}, can be used to model normal data. Especially, AEs have been applied widely as a state-of-the-art approach in network intrusion detection due to the power of modeling complex characteristics of network data \citep{sae-model, nbaiotdataset, Fed-ANIDS}.
By modeling the normal data, new data points will be classified based on the reconstruction error of these ones, a higher error suggesting a higher probability of being an anomaly. 

One of the fundamental challenges in FL is the handling of non-Independent and Identically Distributed (non-IID) data \citep{FL-proposal}. In traditional centralized machine learning, data is typically assumed to be Independent and Identically Distributed (IID), which means that each data point is independent of the others and is drawn from the same distribution. In FL, however, the data on each client device is inherently linked to the user’s behavior and usage patterns. Each device operates within its unique local environment, causing its generated data to have different statistical characteristics, resulting in heterogeneity in IoT networks \citep{noniid-issue}.
An IoT network can be divided into multiple subnetworks, each managed by an IoT gateway, also known as an edge server, and located in a different area. The training data size and distribution of each subnetwork typically vary significantly due to differences in network traffic generated by their specific local environments and applications.
This heterogeneity can lead to local models that are poorly representative of the overall population, making the global model work ineffectively in intrusion detection. 
Addressing this issue requires careful consideration of how to weigh contributions from different gateways to ensure that the global model remains fair and representative of all gateways. 
Both FedAvg and FedProx, mentioned in Section \ref{sec:background}, update the global model based on the size of the training data held by local clients. This makes them face challenges in complex distributed networks with dynamic environments where participant availability and data characteristics can vary significantly. In this case, the client that has a large number of data samples may not represent strongly for the whole network.

In federated IoT applications, it is also crucial to consider some resource constraints aspects such as communication overhead, and computational expense \citep{FL-proposal, noniid-issue}.
Distributed training requires a massive number of clients to participate in the training process, increasing the communication throughput and time consumption. IoT devices have small computational capacity and available resources for training and predicting simultaneously are limited. More accurate approaches require more complex computational activities, leading to an increase in resource consumption and incident response time. Therefore, choosing a lightweight approach and robust learning strategy is essential in federated IoT intrusion detection.
In some research, AEs were used as a feature manipulation method to extract the most important patterns and compress high-dimensional data to more optimal space, such as \cite{sae-model}, \cite{VQuanNguyen-DNACE}. This not only helps reduce the computational complexity but also increases the accuracy and the response time to incidents.

The first research applying FL in IoT intrusion detection is Di\"oT \citep{diot}, a device-type-specific self-learning framework with a GRU network. It consists of two main components: The security gateway acts as an access point for IoT devices and the IoT security service maintains a set of device-type-specific models.
The system autonomously learns and updates its models without requiring human intervention or labeled data. This allows DIoT to adapt to new device behaviors and emerging threats dynamically. The authors conducted extensive experiments using over 30 IoT devices with Mirai botnet in both laboratory and real-world smart home deployment settings. The evaluation showed that DIoT achieved a 95.6\% detection rate with zero false alarms in real-world conditions. 
However, each device type in the system maintains its model on gateways, which can make system management challenging as the scale increases. 
With heterogeneous IoT networks with various device types and huge amounts of data, IoT gates may suffer from computational overload, affecting the response activities.
This research is also limited to the Mirai botnet threats.

\citeauthor{gru-ensemble} utilized LSTM and GRU with the ensemble learning technique to enable on-device learning. The predictions from local GRU models are combined using a Random Forest Classifier (RFC). Each GRU model provides probability values for each potential label (attack type) for a given input. The RFC aggregates the probability values from all GRU models. It uses these probabilities as votes to determine the final prediction. The label with the highest combined probability across all models is selected as the final prediction. Their evaluation demonstrated the performance of the FL approach compared to the non-FL approach with a high anomaly detection rate of 99.5\% and a minimal number of false alarms.
However, the ability of this approach to detect unknown intrusion is limited due to the supervised learning technique on available attacks. 
The proposed ensemble approach may be impacted by the heterogeneity in modern IoT networks and the communication overhead in the prediction phase when requiring all clients to infer new data, and not considering the characteristics of local data.

Fed-ANIDS \citep{Fed-ANIDS} is a network-based intrusion detection approach based on the Autoencoder model similar to our proposal, a semi-supervised learning method. Various autoencoder-based models, including simple Autoencoders (AE), Variational Autoencoders (VAE), and Adversarial Autoencoders (AAE), are utilized for anomaly detection based on reconstruction errors of normal traffic. The FL setting was employed by using FedAvg and FedProx algorithms. The authors conducted several experiments on some well-known network traffic datasets such as USTC-TFC2016, CIC-IDS2017, and CSE-CIC-IDS2018. 
The results demonstrated that Fed-ANIDS achieves high accuracy in detecting network intrusions while maintaining low false alarm rates and preserving privacy. Additionally, they show that FedProx slightly outperforms FedAvg in terms of accuracy. Nevertheless, the findings also highlight the impact of heterogeneous network settings. This approach faces challenges in effectively generalizing to normal network data and performs poorly on unseen data.

%% file: Sections/4-Methodology.tex
\section{Proposed approach}
\label{sec:proposed-approach}

\begin{figure*}[t]
  \centering
    \includegraphics[width=\textwidth,height=0.96\textheight]{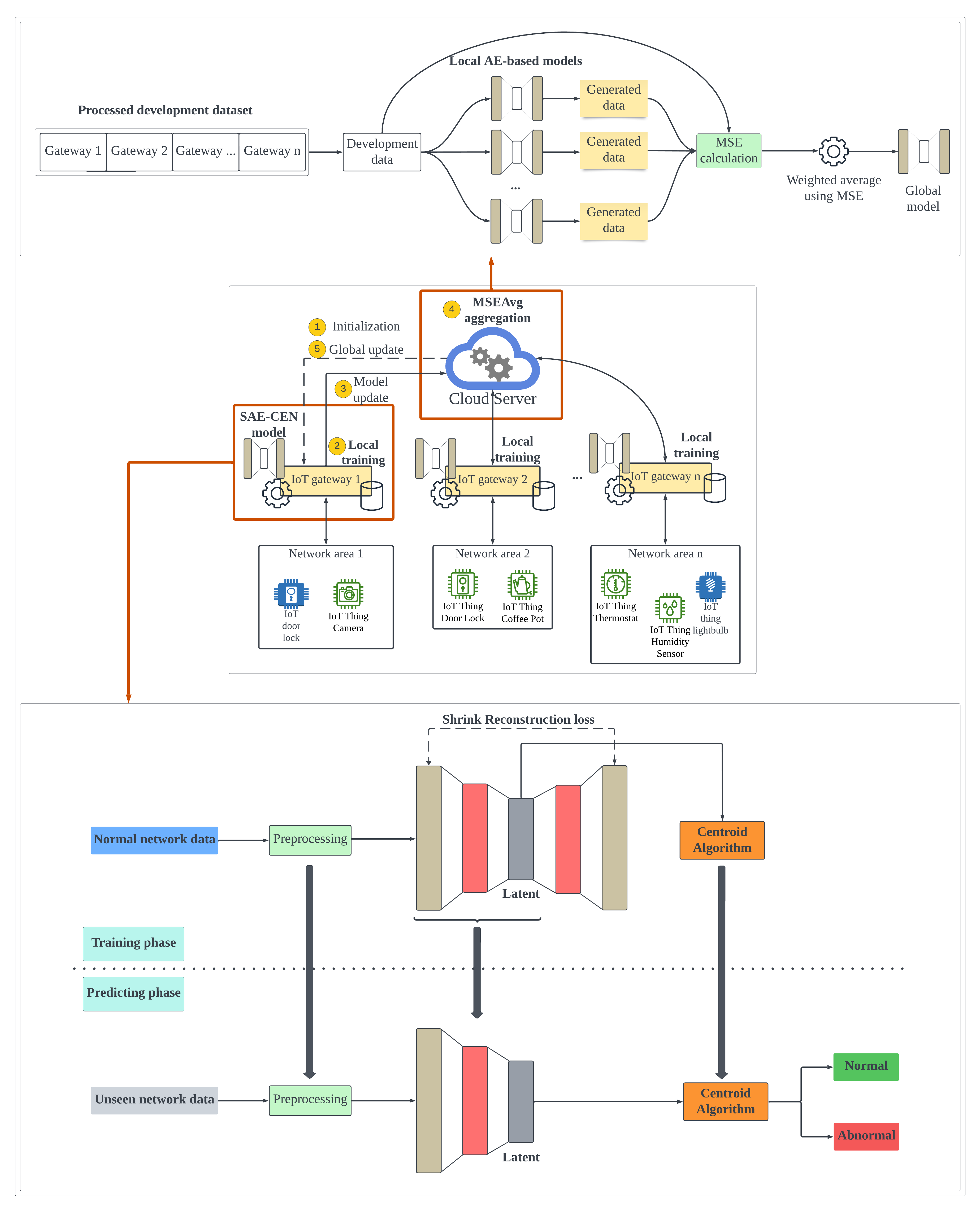} 
  \caption{Overview of our proposed approach.}
  \label{fig:overview-proposed}
\end{figure*}

In this paper, we propose \textbf{FedMSE}, an IoT intrusion detection approach using federated learning and a hybrid model based on the Shrink Autoencoder (an Autoencoder variant) and the Centroid algorithm. Figure \ref{fig:overview-proposed} presents the overall architecture. It consists of two main components as described in the federated learning scheme: \textit{(1) SAE-CEN hybrid intrusion detector}, \textit{(2) MSEAvg aggregation}. The details of these two parts are explained in the rest of this section.

To feed to the machine learning model, firstly, the network data needs to be preprocessed. Network traffic is captured and extracted to tabular features as proposed in \citeauthor{nbaiotdataset}. We use the Standard Normalization method to normalize the data to reduce the computational complexity and ensure that all features contribute equally to the training process. The formula for standard scaling (also known as Z-score normalization) is given by:
\begin{equation}
    z = \frac{x - \mu}{\sigma}
\end{equation}
where \( x \) is the original feature value, \( \mu \) is the mean of the feature, \( \sigma \) is the standard deviation of the feature.

\subsection{Hybrid intrusion detection approach}
By modelling normal network data, an Autoencoder-based model can take advantage of its latent layer to transform the original data into a new data space where the data has a smaller number of dimensions and presents the most important characteristics. Hence, some common traditional one-class classifiers can work effectively on this new data to detect anomalies. In this part, a hybrid model is constructed using Shrink Autoencoder (SAE) as a data representation method and Centroid anomaly detection algorithm (CEN) as a detector. This approach is named the SAE-CEN model.

Shrink Autoencoder (SAE) \citep{sae-model} is a powerful data representation model that helps common network intrusion detection algorithms deal with sparse and high-dimensional network data, even with a small amount of training data. This model is an Autoencoder variant since a new regularization term is added to the Autoencoder objective function in Equation \ref{eq:ae-loss}. The new loss function is formulated as Equation \ref{eq:sae}. This makes it easier to construct the normal network data in the latent layer. 

The objective function of the SAE training process is formulated as follows:
\begin{equation}
\label{eq:sae}
\mathcal{L}_{\text{SAE}}(\lambda; x^i, \hat{x}^i, h) = \frac{1}{n} \sum_{i=1} \| x^i - \hat{x}^i \|^2 + \lambda \frac{1}{n} \sum_{i=1}^{n} \|h^i\|^2
\end{equation}
where $\hat{x}^i$ and $h^i$ are the reconstruction output and the latent vector of the input $x_i$, correspondingly. The parameter $\lambda$ controls the trade-off between the two terms in the equation. It is optimized using the Algorithm \ref{alg:local-training}. 

\begin{algorithm}[ht]
\caption{GatewayUpdate: Optimize local models}
\label{alg:local-training}
\begin{algorithmic}[1]
\REQUIRE Local normal data $X = [x_0, ..., x_{M-1}] \sim p(x)$; Number of local epochs $I$; Set of mini-batches $\mathcal{B}$ each of size $N$; Learning rate $\eta$; Shrink regularizer factor $\lambda$;
\ENSURE Trained Shrink Autoencoder

\STATE Receive initial weights $\theta_e$ for encoder and $\theta_d$ for decoder from server
\FOR{epoch = 1 to $I$}
    \FOR{each mini-batch $x_i$ in $X$}
        \STATE $h_i \leftarrow \text{Encoder}(x_i; \theta_e)$
        \STATE $\hat{x}_i \leftarrow \text{Decoder}(h_i; \theta_d)$
        \STATE $\mathcal{L}_{\text{SAE}} \leftarrow \frac{1}{N} \sum_{i=0}^{N-1} \|x_i - \hat{x}_i\|^2 + \lambda \frac{1}{N} \sum_{i=0}^{N-1} \|h_i\|^2$
        \STATE Compute gradients $\nabla_{\theta_e} \mathcal{L}_{\text{SAE}}$ and $\nabla_{\theta_d} \mathcal{L}_{\text{SAE}}$
        \STATE $\theta_e \leftarrow \theta_e - \eta \nabla_{\theta_e} \mathcal{L}_{\text{SAE}}$
        \STATE $\theta_d \leftarrow \theta_d - \eta \nabla_{\theta_d} \mathcal{L}_{\text{SAE}}$
    \ENDFOR
\ENDFOR
\STATE Send $\theta_e$, $\theta_d$ to the global server
\end{algorithmic}
\end{algorithm}

Given a normal network data set $X_0 = \{x_0^{(1)}, \ldots, x_0^{(n)}\} \subset \mathbb{R}^d$, the goal of intrusion detection is to determine whether a new data point $\mathbf{x}$ has the same probability characteristics as the set $X_0$. 
A straightforward approach to anomaly detection involves estimating the probability density function (PDF) of the distribution from which the dataset $X_0$ is derived. 
If a new instance $\mathbf{x}$ is located in an area of the distribution where the density is low, it is flagged as anomalous. 
However, estimating the density of a distribution is a challenging task, particularly in high-dimensional spaces. 
Another simple way comes from clustering-based algorithms with the assumption that normal data instances lie close to their cluster centroid. 
Meanwhile, abnormal data points are far from the centroid~\citep{anomaly-detection-survey}, such as the Centroid (CEN) algorithm. 
The central idea is to leverage the distance from an observation to the centroid of the cluster as the abnormality of the observation. 
This distance is also known as the anomaly score. A higher score suggests that the data point has a higher probability of being an anomaly. 
By specifying a threshold, a query data point can be classified as normal or anomalous. 
The CEN algorithm is a very simple algorithm with no hyper-parameters, and its computational expense is small. 

\begin{figure}[ht]
  \centering
  \includegraphics[width=\linewidth,height=0.2\textheight]{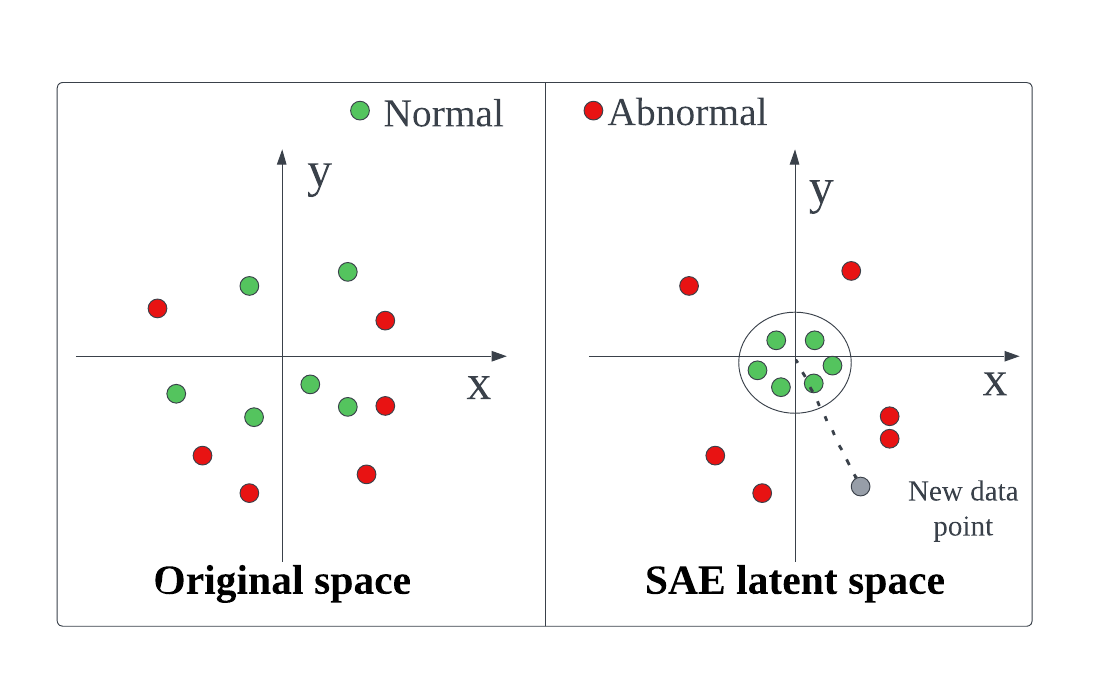} 
  \caption{Illustration of SAE-CEN hybrid model principle.}
  \label{fig:sae-cen-prin}
\end{figure}

After receiving the best SAE model from the server, the decoder part will be removed, and the encoder will be used as a data manipulator that forms the data in good shape to help CEN work more effectively (Figure \ref{fig:sae-cen-prin}). 
This also reduces the time required to respond to incidents. Therefore, the SAE-CEN combination can not only work powerfully to detect anomalies but also respond to incidents rapidly. 

\cite{sae-model} demonstrated that the CEN performed very accurately under the SAE representation data in centralized network anomaly detection. In this research, we used a federated learning scheme to evaluate the accuracy of this solution in the decentralized IoT intrusion detection task.

\subsection{Mean Squared Error-based model aggregation}
The core component in all federated learning architectures is the global model aggregation. How effectively the global model is updated decides the power of the FL scheme. This section provides the design of a novel FL algorithm that boosts the performance of an Autoencoder-based model, especially SAE-CEN.

During training a machine learning model, it is necessary to control the convergence of the model on both local and global sides to prevent under-fitting and over-fitting problems. 
In the data preprocessing phase, the server will collect a small processed dataset from all gateways in the network, called development data, that includes normal network data only to validate the global model convergence in each training round. 
Each gateway randomly selects the same amount of normal data as the others, and these subsets are then combined to form the development dataset,
ensuring the representativeness of the whole IoT network. Using processed data also keeps the sensitive information private in local gateways.

This research leverages as much as possible the strength of the Autoencoder-based model in modelling normal data and proposes the \textbf{MSEAvg} algorithm to aggregate the global model by comparing the ability of each local model in reconstructing the above development dataset. The model that works better will play a more critical role in updating the global model.

Figure \ref{fig:overview-proposed} and Algorithm \ref{alg:mseavg} show the MSEAvg principle and pseudo-code for the whole operation. 
When all participants have completed the local training process and sent the optimal weights to the server, each local model uses the development data as input, returns the reconstruction error calculated by Mean Squared Error (MSE) of input and output, and is assigned a weight in the updating process based on this error. 
The smaller error implies a better model for learning normal data, which means a higher weight.
After that, the global model will be aggregated by using the Equation \ref{eq:mseavg}:

\begin{equation}
\label{eq:mseavg}
    W_{global} = \frac{\sum_{i=1}^n \alpha_i \cdot W_i}{\sum_{i=1}^n \alpha_i}
\end{equation}
where $W_i$ is the local model weights; $\alpha_i$ is the update weight based on MSE loss and calculated by Equation $\ref{eq:alpha_i}$:

\begin{equation}
    \label{eq:alpha_i}
    \alpha_i = \frac{1}{MSE_i}
\end{equation}
where $MSE_i$ is the mean squared error of local model $i$ with the development dataset.

\begin{algorithm}[H]
\caption{GlobalAgg: MSEAvg aggregation}
\label{alg:mseavg}
\begin{algorithmic}[1]
\REQUIRE Number of global round $E$; Local models $L = \{L_1, L_2, \ldots, L_n\}$; Development dataset $D$;
\ENSURE Updated global model $M$

\STATE Initialize list $update\_weights \leftarrow []$
\FOR{each local model $L_i$ in $L$}
    \STATE Generate new development data: $\hat{D_i} \leftarrow L_i(D)$
    \STATE Calculate similarity score between $D$ and $\hat{D}$: $sim\_score \leftarrow Mean\_Squared\_Error(D, \hat{D})$ 
    \STATE Compute update weight $w_i \leftarrow 1 / sim\_score$
    \STATE Append $(L_i, w_i)$ to $update\_weights$
\ENDFOR
\FOR{each ($L_i$, $\alpha$) in $update\_weight$}
    \STATE Compute $M \leftarrow \frac{\sum L_i * \alpha}{\sum_{\alpha}}$
\ENDFOR
\STATE Send updated model $M$ to all gateways for detection

\end{algorithmic}
\end{algorithm}

The MSEAvg algorithm enhances the global federated Autoencoder-based models by prioritizing contributions from local models with superior reconstruction accuracy, as models with lower reconstruction errors on the development dataset better represent normal network data. 
This approach ensures a robust and reliable global model by aligning the aggregation process with the primary objective of Autoencoder-based models in anomaly detection.
In addition, MSE is more sensitive in model aggregation than alternatives such as Mean Absolute Error (MAE), which helps reduce the effect of low-performing local models. MSE squares the reconstruction errors, which amplifies the effect of larger errors during the weight calculation more than MAE. Poor-performing local models with high reconstruction errors are assigned much smaller weights. Meanwhile, well-performing local models with lower errors contribute more significantly to the global model. This helps the global model be more robust.

Unlike conventional algorithms such as FedAvg and FedProx, which weights contributions based on dataset size, MSEAvg employs a performance-based criterion. 
This innovation is particularly effective in handling non-IID data. 
By evaluating local models using a representative development dataset, MSEAvg ensures fair assessment regardless of training data heterogeneity. 
This performance-based weighting reduces the risk of over-fitting to data-rich gateways, thereby enhancing the generalizability and reliability of the global model.

%% file: Sections/5-0-Evaluation.tex
\section{Evaluation}
\label{sec:evaluation}
\input{Sections/5-1-Settings}
\input{Sections/5-2-Result-Discussion}

%% file: Sections/5-1-Settings.tex
This section presents the evaluation of our approach and focuses on resolving three research questions that correspond to the three experiments below.
\begin{enumerate}
    \item \label{rq1} \textbf{RQ1}. \textit{How does the proposed approach work in IoT federated intrusion detection?}
    
    We perform Experiment 1 to solve this RQ. Section \ref{exp:exp1}  presents the results of this experiment and discusses some findings to examine the ability of our approach in federated IoT network intrusion detection. 
    \item \label{rq2} \textbf{RQ2}. \textit{How does the gateway selection ratio affect federated learning intrusion detection?}

    Conducting Experiment 2 corresponds to solving RQ2. Section \ref{exp:exp2} shows the results of Experiment 2 to consider the efficiency of our approach with various gateway selection ratios.
    \item \label{rq3} \textbf{RQ3}. \textit{How does the proposed approach work on large-scale IoT networks?}

    The robustness of the proposed approach in large-scale IoT networks will be investigated with the findings of Experiment 3 in Section \ref{exp:exp3}.
    
\end{enumerate}

\subsection{Experiment settings}
Below we present the settings used for the experiments we conducted in order to evaluate our approach. The actual experiments will be discussed in Sections \ref{exp:exp1} through \ref{exp:exp3}.

\subsubsection{IoT network construction}
Relying on the description of non-IID challenges in FL, we conduct experiments in two scenarios, 1) \textit{low non-IIDness} and 2) \textit{high non-IIDness}, to evaluate our approach.
For the first context, we consider an IoT network in which the gateways have the same topology, device type, and applications, resulting in an equivalent data distribution across these subnetworks. 
In contrast, the second setting abstracts an IoT network in which subnetworks differ in all topologies, IoT device types, applications, and the number of data records for each device type.

To mimic the IoT network for this evaluation, a small partition of the N-BaIoT dataset, with only normal data, is used to make experiments close to practical scenarios. N-BaIoT dataset \citep{nbaiotdataset} is designed to aid in the detection of IoT botnet attacks. The dataset was collected from nine commercial IoT devices that were intentionally infected with two prevalent IoT-based botnets: Mirai and Gafgyt. The details of this dataset are described in Table \ref{tab:nbaiotdataset}.

\begin{table*}[h]
    \caption{N-BaIoT network traffic dataset. \textbf{Normal}, \textbf{Gafgyt} and \textbf{Mirai} columns show the number of data samples for normal, Gafgyt botnet and Mirai botnet behaviour in each IoT device, respectively.}
    \centering
    \label{tab:nbaiotdataset}
    \begin{tabular}{|c|l|c|c|c|}
        \hline
        \textbf{ID} & \multicolumn{1}{c|}{\textbf{Device name}}     & \textbf{Normal} & \textbf{Gafgyt} & \textbf{Mirai} \\ \hline
        0           & Danmini\_Doorbell                             & 49548           & 652100          & 316650         \\ \hline
        1           & Ecobee\_Thermostat                            & 13113           & 512133          & 310630         \\ \hline
        2           & Philips\_B120N10\_Baby\_Monitor               & 175240          & 312273          & 610714         \\ \hline
        3           & Provision\_PT\_737E\_Security\_Camera         & 62154           & 330096          &                \\ \hline
        4           & Provision\_PT\_838\_Security\_Camera          & 98514           & 309040          & 429337         \\ \hline
        5           & Samsung\_SNH\_1011\_N\_Webcam                 & 52150           & 323072          &                \\ \hline
        6           & SimpleHome\_XCS7\_1002\_WHT\_Security\_Camera & 46585           & 303223          & 513248         \\ \hline
        7           & SimpleHome\_XCS7\_1003\_WHT\_Security\_Camera & 19528           & 316438          & 514860         \\ \hline
        8           & Ennio\_Doorbell                               & 39100           & 316400          &                \\ \hline
    \end{tabular}
\end{table*}

\begin{figure*}[h]
  \centering

  \begin{subfigure}[b]{0.49\textwidth}
    \centering
    \includegraphics[width=\textwidth, height=0.27\textheight]{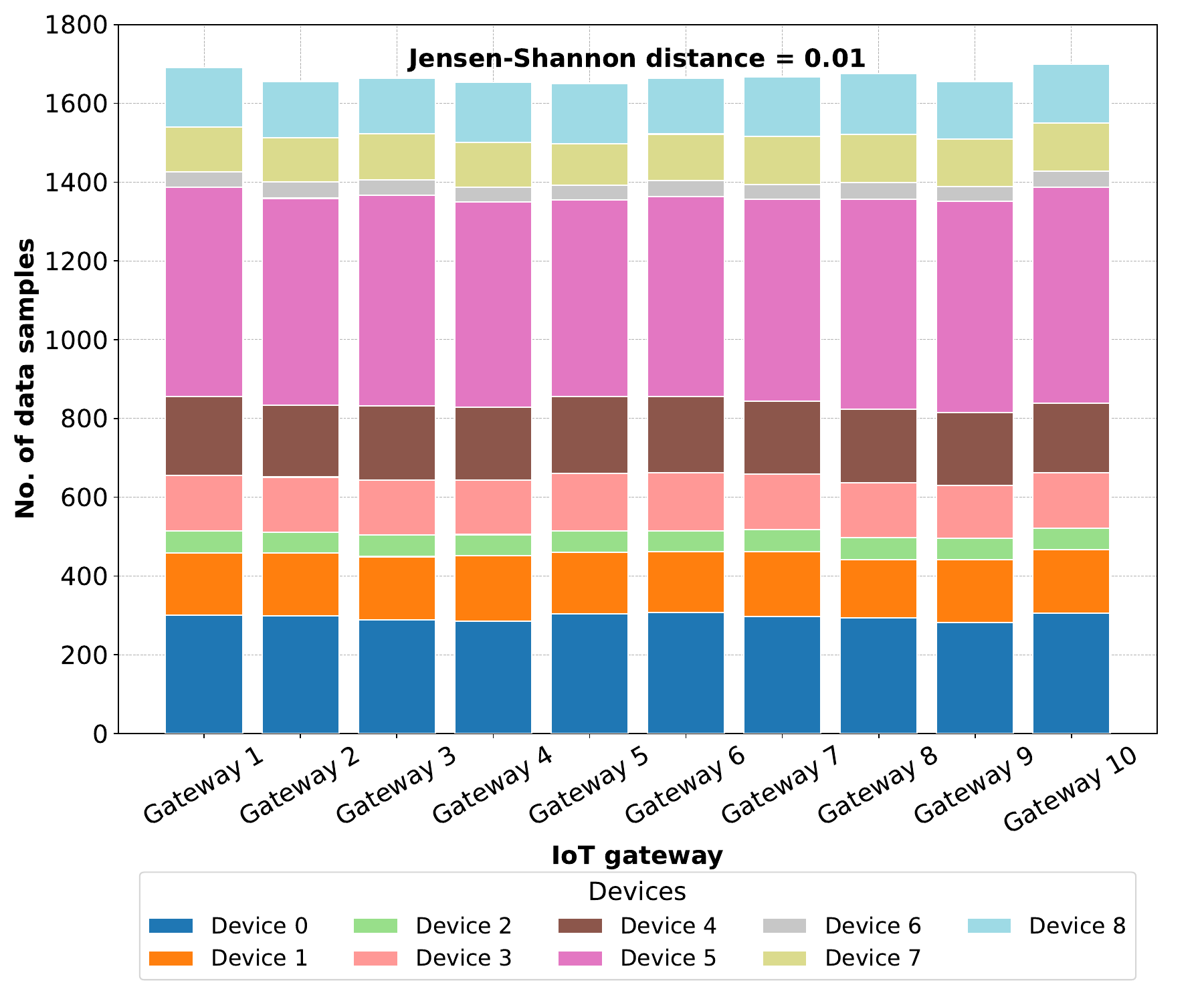} 
    \captionsetup{justification=centering}
    \caption{Training data with low non-IIDness}
    \label{fig:iid-dataset-training}
  \end{subfigure}
  \hfill
  \begin{subfigure}[b]{0.49\textwidth}
    \centering
    \includegraphics[width=\textwidth, height=0.27\textheight]{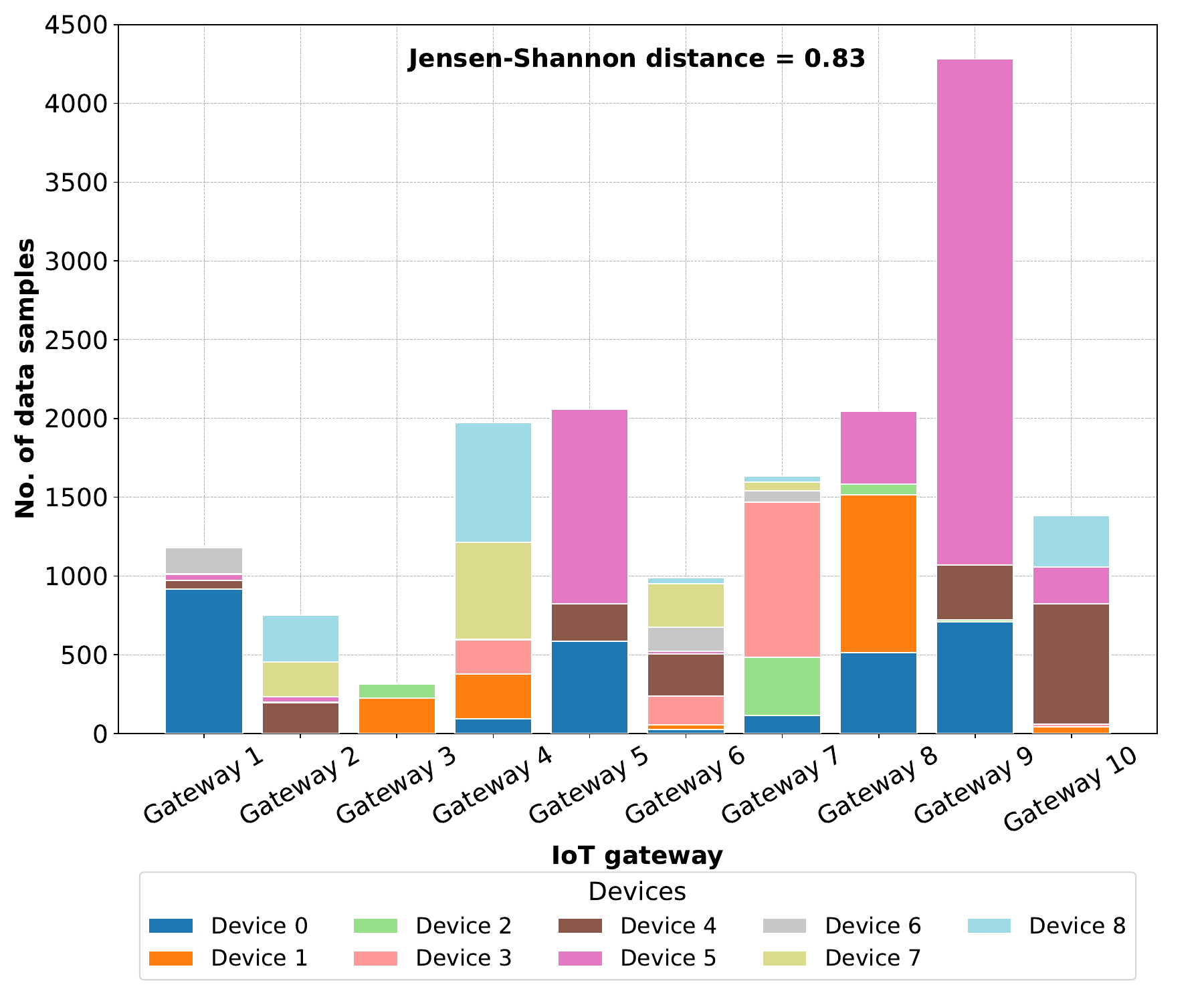} 
    \captionsetup{justification=centering}
    \caption{Training data with high non-IIDness}
    \label{fig:non-iid-dataset-training}
  \end{subfigure}

  \vspace{0.5em} 
  \begin{subfigure}[b]{0.49\textwidth}
    \centering
    \includegraphics[width=\textwidth, height=0.27\textheight]{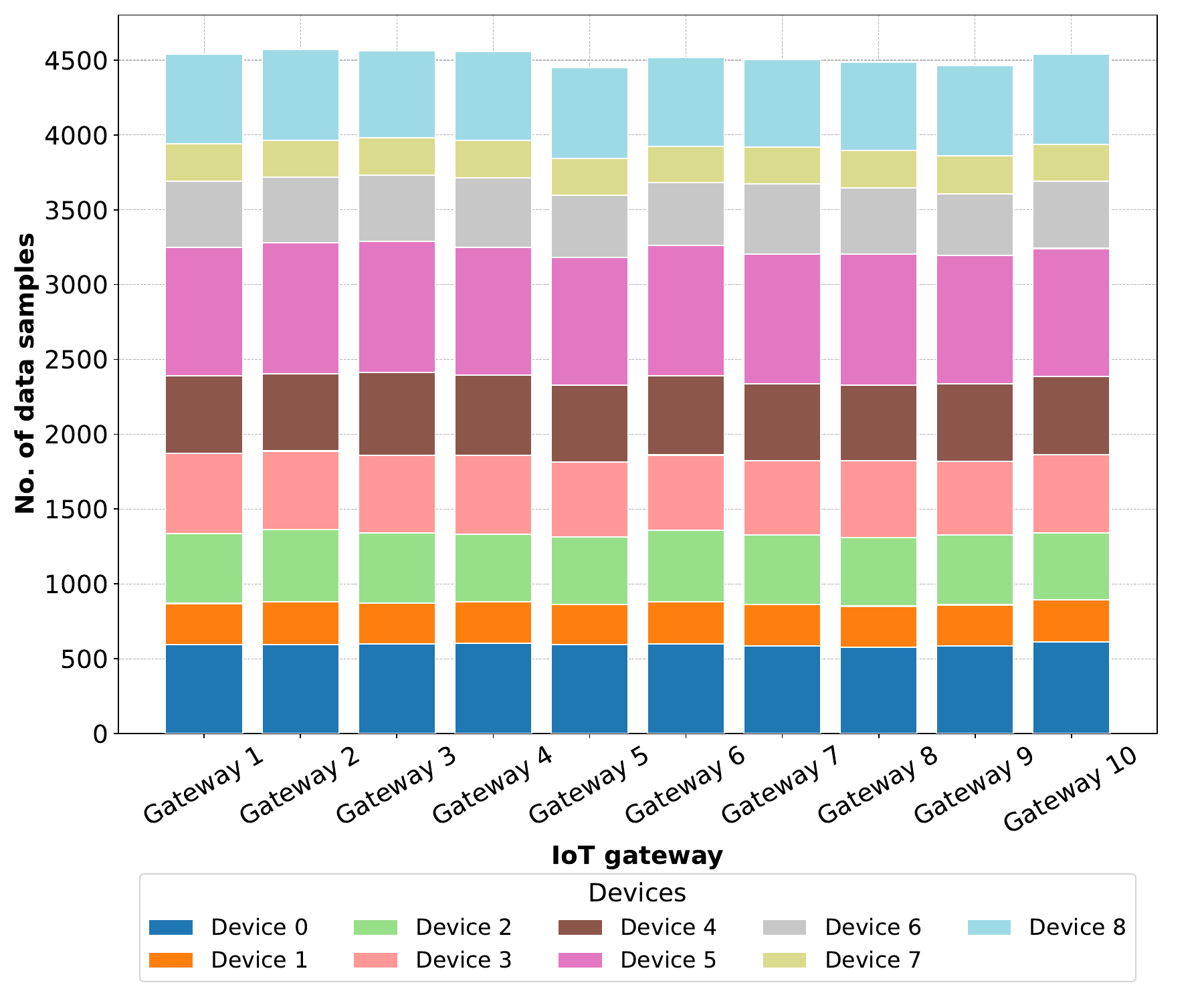} 
    \captionsetup{justification=centering}
    \caption{Testing data with low non-IIDness}
    \label{fig:iid-dataset-testing}
  \end{subfigure}
  \hfill
  \begin{subfigure}[b]{0.49\textwidth}
    \centering
    \includegraphics[width=\textwidth, height=0.27\textheight]{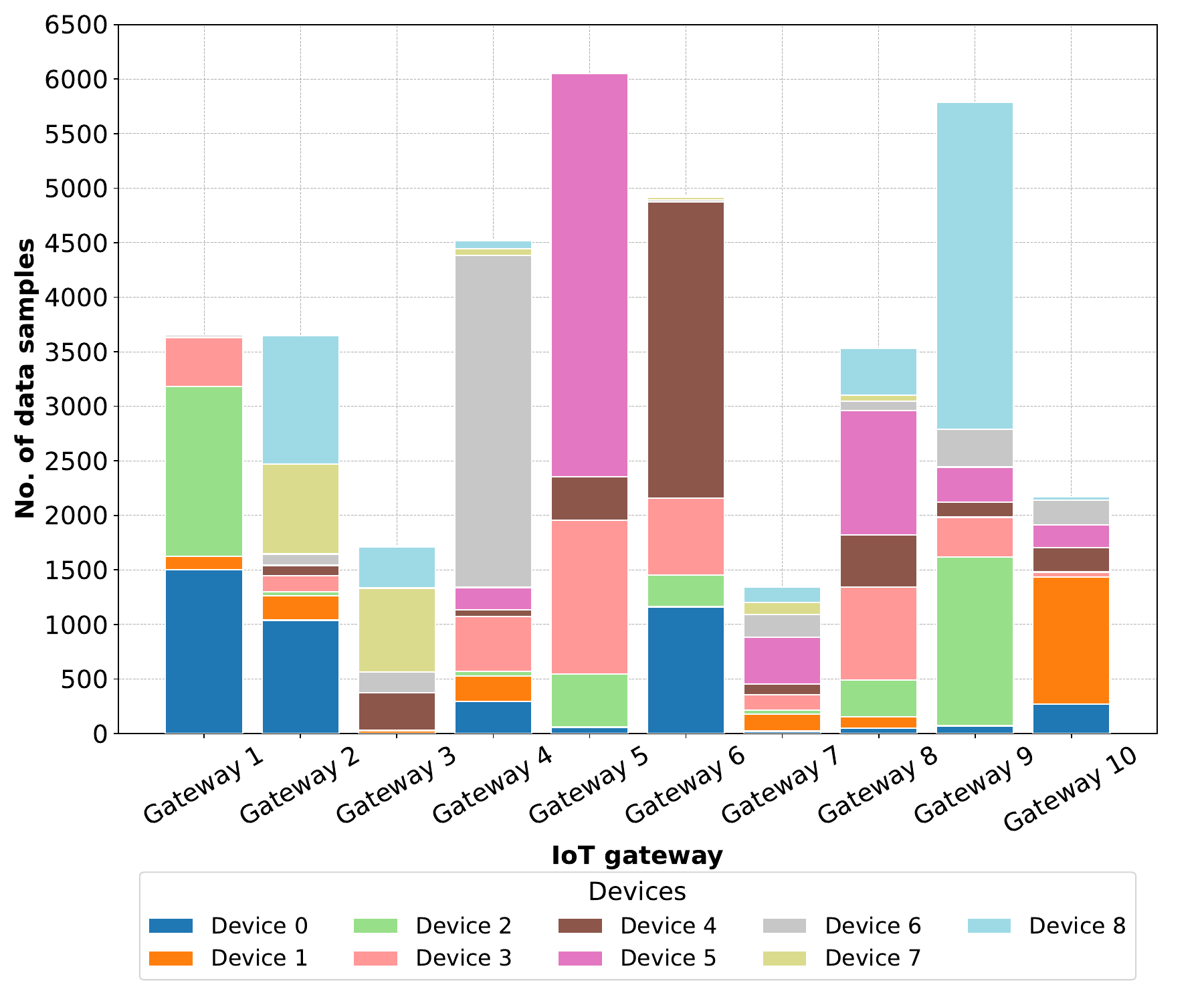} 
    \captionsetup{justification=centering}
    \caption{Testing data with high non-IIDness}
    \label{fig:noniid-dataset-testing}
  \end{subfigure}

  \caption{Data allocation for experimental scenarios. The training data (a, b) contains only normal data. The testing data (c, d) includes both normal data and abnormal data and some new IoT devices appear in subnetworks.}
  \label{fig:combined-dataset}
\end{figure*}

Figure \ref{fig:combined-dataset} shows the experimental data allocation. We consider a 10-gateway IoT network, which means that there are 10 IoT gateways in the network. 
The gateway data is selected from N-BaIoT data using random algorithms and constraints to ensure the non-IID characteristics of training data. 
Following the approach in \cite{fedart-ml}, we use the Dirichlet distribution $Dir_{n}(\alpha)$ with $n$ as the number of gateways and $\alpha$ as the concentration parameter and Jensen-Shannon measurement $(\textit{JS})$ to control the non-IIDness of networks.
To simulate a homogeneous setting, for each device type $k$, we sample $p_k \sim Dir_{10}(1000)$ and allocate the proportion data $p_{k,j}$ of device $k$ for the gateway $j$, and the measurement is $\text{\textit{JS}} = 0.01$. 
We do the same process for the heterogeneous setting with the concentration parameter $\alpha = 0.1995$ and $\text{\textit{JS}} = 0.83$. 
In the testing phase, some new devices will be added to the network to simulate the dynamics of IoT networks.

To solve the \textbf{RQ3} when considering IoT networks in different network scales, we use the same methodology to select the data in both those contexts for 20-gateway, 30-gateway, 40-gateway, and 50-gateway networks that have 20 gateways, 30 gateways, 40 gateways and 50 gateways in the topology, correspondingly.

\subsubsection{Hyperparameters setting}

This research studies the accuracy of Autoencoder and SAE-CEN models under FedAvg, FedProx, and MSEAvg federated learning algorithms. In intrusion detection, it is challenging to determine a threshold that accurately classifies a data point as either normal or anomalous. Thus, the Area Under the ROC Curve (AUC) metric \citep{AUC} is adopted to evaluate these models at different thresholds. We ran each experiment five times and calculated the mean accuracy and standard deviation of the AUC values, following the research in \cite{FL-malware-iot} that evaluated on N-BaIoT dataset.

Due to the lack of anomalies during training, the experiment hyperparameters can not be tuned, which is one of the considerable challenges of this work. We set up them using common values. The mini-batch size value is chosen as $12$, learning rate $lr = 0.00001$, local epoch $I = 100$, number of global round $E = 20$, $\mu = 0.001$ for the FedProx proximal term. In each round, we randomly select half of the gateways to join the training process based on the majority rule. For the SAE shrink parameter, we set the value of 10 as \cite{sae-model} for balancing shrink loss and MSE loss and employ the Adam optimizer \citep{adamoptimizer} along with early stopping techniques \citep{earlystopping} to train these local models. 
We split the local training data into 40\% for training, 10\% for validation, 40\% for the development dataset, and 10\% for adding to testing set. To ensure all gateways have an equal amount of development data, we use the size of the gateway that has the minimum number of development data samples.
The early stopping techniques are also used in the global server to control the convergence of the global model on the development dataset. For the number of neurons in the latent layer of the Autoencoder-based model, we configure based on rules of thumb as mentioned in \cite{ruleofthump} with the value \( m = \left[ 1 + \sqrt{n} \right] \), where $n$ is the original space dimension. 

All experiments were implemented in Python language and run on one KAGAYAKI high-performance computing GPU server at Japan Advanced Institute of Science and Technology, which has an Intel Xeon GOLD 5320 52-core 2.2 GHz CPU, 512 GB DDR4/3200 SDRAM memory, and two NVIDIA A100 48 GB GPUs.

%% file: Sections/5-2-Result-Discussion.tex
\subsection{Experiment 1: Federated intrusion detection performance}
\label{exp:exp1}

\begin{table*}[!ht]
\centering
\caption{AUCs (\%) for different gateways in low non-IIDness and high non-IIDness settings. FedMSE results are represented by the combination of SAE-CEN with the MSEAvg algorithm.}
\label{tab:exp1-result}
\resizebox{\textwidth}{!}{%
\begin{tabular}{@{}ccccccccccc@{}}
\toprule
\multirow{2}{*}{\textbf{Agent}} & \multicolumn{3}{c}{\textbf{Autoencoder}} & \multicolumn{3}{c}{\textbf{SAE-CEN}} \\ \cmidrule(lr){2-4} \cmidrule(lr){5-7}
& \textbf{FedAvg} & \textbf{FedProx} & \textbf{MSEAvg} & \textbf{FedAvg} & \textbf{FedProx} & \textbf{MSEAvg} \\ \midrule
\multicolumn{7}{c}{\textit{a) Low non-IIDness}} \\ \midrule
\textbf{Gateway 1} & 98.77$\pm$0.33 & 98.53$\pm$0.51 & 98.54$\pm$0.42 & 97.93$\pm$0.75 & 97.67$\pm$1.18 & 98.68$\pm$0.64 \\
\textbf{Gateway 2} & 97.81$\pm$0.50 & 97.53$\pm$0.59 & 97.51$\pm$0.73 & 97.73$\pm$0.75 & 97.76$\pm$0.73 & 97.83$\pm$0.82 \\
\textbf{Gateway 3} & 98.43$\pm$0.47 & 98.10$\pm$0.57 & 98.02$\pm$0.58 & 97.80$\pm$0.55 & 97.74$\pm$0.49 & 98.26$\pm$0.85 \\
\textbf{Gateway 4} & 99.42$\pm$0.20 & 99.39$\pm$0.25 & 99.32$\pm$0.30 & 99.22$\pm$1.00 & 98.52$\pm$0.82 & 99.31$\pm$0.70 \\
\textbf{Gateway 5} & 99.75$\pm$0.10 & 99.75$\pm$0.04 & 99.73$\pm$0.07 & 99.84$\pm$0.06 & 99.81$\pm$0.07 & 99.86$\pm$0.05 \\
\textbf{Gateway 6} & 99.23$\pm$0.26 & 99.16$\pm$0.31 & 99.09$\pm$0.35 & 98.92$\pm$0.92 & 98.22$\pm$0.69 & 99.16$\pm$0.41 \\
\textbf{Gateway 7} & 98.30$\pm$0.35 & 98.08$\pm$0.45 & 98.06$\pm$0.39 & 97.82$\pm$0.78 & 97.59$\pm$1.28 & 98.30$\pm$0.79 \\
\textbf{Gateway 8} & 100.00$\pm$0.00 & 100.00$\pm$0.00 & 100.00$\pm$0.00 & 100.00$\pm$0.00 & 100.00$\pm$0.00 & 100.00$\pm$0.00 \\
\textbf{Gateway 9} & 99.03$\pm$0.17 & 98.94$\pm$0.17 & 98.93$\pm$0.20 & 98.38$\pm$0.76 & 98.06$\pm$1.22 & 98.72$\pm$0.54 \\
\textbf{Gateway 10} & 99.97$\pm$0.02 & 99.97$\pm$0.02 & 99.97$\pm$0.02 & 99.95$\pm$0.02 & 99.95$\pm$0.03 & 99.95$\pm$0.02 \\
\textit{\textbf{Average}} & \textit{\textbf{99.07$\pm$0.24}} & \textit{98.95$\pm$0.29} & \textit{98.92$\pm$0.31} & \textit{98.76$\pm$0.56} & \textit{98.53$\pm$0.65} & \textit{99.01$\pm$0.48} \\ \midrule
\multicolumn{7}{c}{\textit{b) High non-IIDness}} \\ \midrule
\textbf{Gateway 1} & 91.00$\pm$4.84 & 87.96$\pm$6.25 & 86.61$\pm$1.92 & 96.65$\pm$1.00 & 97.44$\pm$1.62 & 97.25$\pm$1.04 \\
\textbf{Gateway 2} & 99.58$\pm$0.03 & 99.58$\pm$0.03 & 99.56$\pm$0.00 & 99.62$\pm$0.05 & 99.61$\pm$0.09 & 99.64$\pm$0.09 \\
\textbf{Gateway 3} & 88.68$\pm$5.76 & 87.70$\pm$5.87 & 86.80$\pm$4.84 & 93.55$\pm$1.09 & 93.82$\pm$1.30 & 94.58$\pm$0.89 \\
\textbf{Gateway 4} & 89.42$\pm$5.14 & 88.72$\pm$4.93 & 86.14$\pm$2.48 & 94.40$\pm$1.11 & 94.41$\pm$1.66 & 94.97$\pm$1.03 \\
\textbf{Gateway 5} & 97.96$\pm$1.19 & 97.53$\pm$1.17 & 96.91$\pm$0.28 & 98.63$\pm$0.47 & 98.76$\pm$0.62 & 98.69$\pm$0.46 \\
\textbf{Gateway 6} & 92.70$\pm$3.52 & 91.75$\pm$3.90 & 89.47$\pm$1.13 & 93.84$\pm$2.03 & 95.38$\pm$0.87 & 94.43$\pm$0.25 \\
\textbf{Gateway 7} & 94.62$\pm$2.81 & 93.94$\pm$3.25 & 92.53$\pm$0.53 & 96.18$\pm$0.34 & 96.63$\pm$0.94 & 96.58$\pm$0.58 \\
\textbf{Gateway 8} & 95.84$\pm$2.14 & 95.47$\pm$2.20 & 94.18$\pm$0.94 & 97.51$\pm$0.55 & 97.87$\pm$0.50 & 97.73$\pm$0.07 \\
\textbf{Gateway 9} & 98.05$\pm$1.39 & 97.65$\pm$1.33 & 97.04$\pm$0.50 & 99.14$\pm$0.33 & 99.02$\pm$0.71 & 99.30$\pm$0.42 \\
\textbf{Gateway 10} & 99.52$\pm$0.09 & 99.52$\pm$0.09 & 99.48$\pm$0.00 & 99.77$\pm$0.05 & 99.81$\pm$0.10 & 99.77$\pm$0.05 \\
\textit{\textbf{Average}} & \textit{94.74$\pm$2.69} & \textit{93.98$\pm$2.90} & \textit{92.87$\pm$1.26} & \textit{96.93$\pm$0.70} & \textit{97.28$\pm$0.84} & \textit{\textbf{97.30$\pm$0.49}} \\ \bottomrule
\end{tabular}%
}
\end{table*}

\begin{figure}[!b]
  \centering
  \includegraphics[width=\columnwidth, height=0.25\textheight]{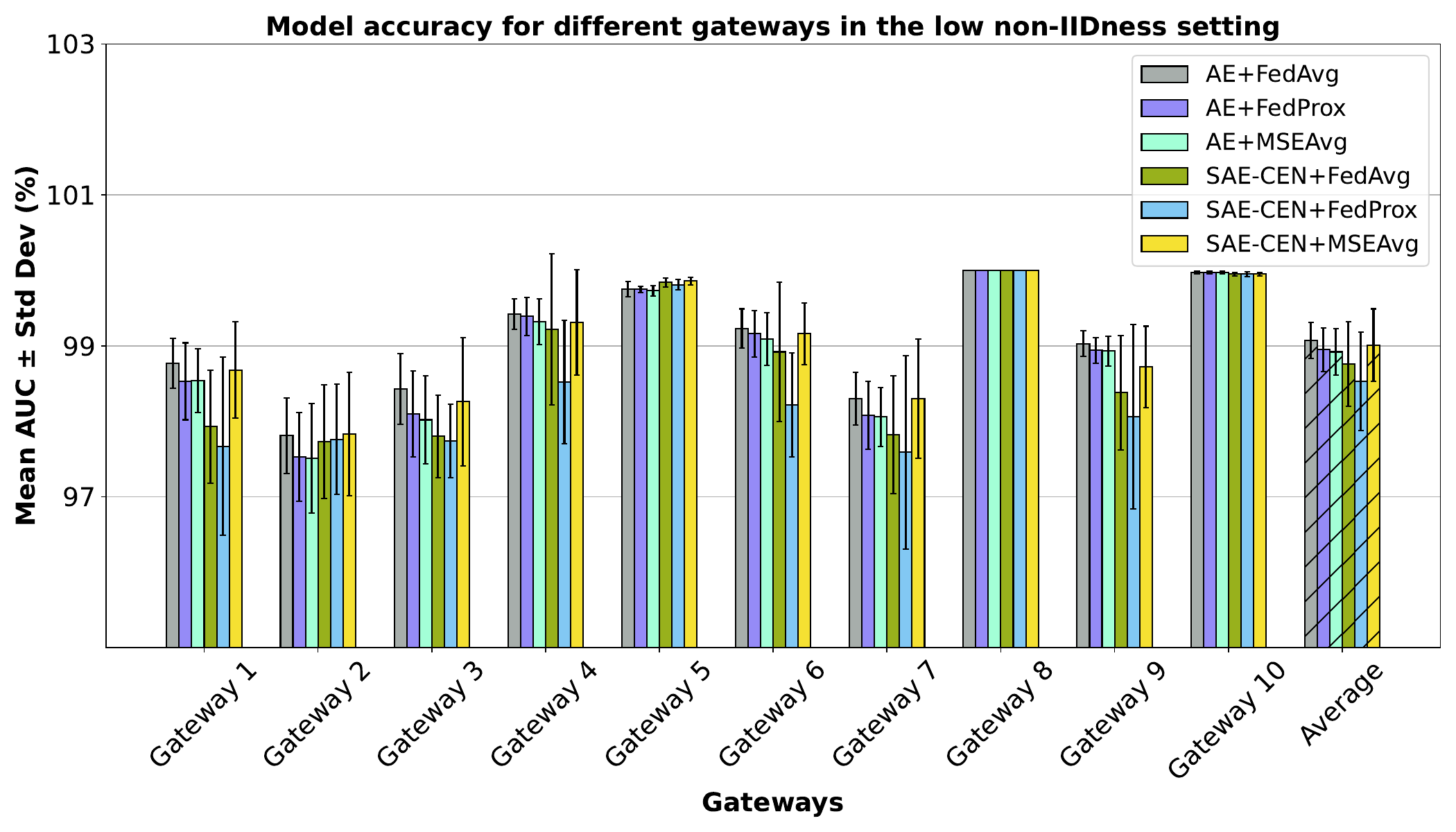}
  \captionsetup{justification=centering} 
  \caption{Accuracy for federated intrusion detection in the low non-IIDness setting}
  \label{fig:iid-exp1}
\end{figure}

\begin{figure}[!b]
  \centering
  \includegraphics[width=\columnwidth, height=0.25\textheight]{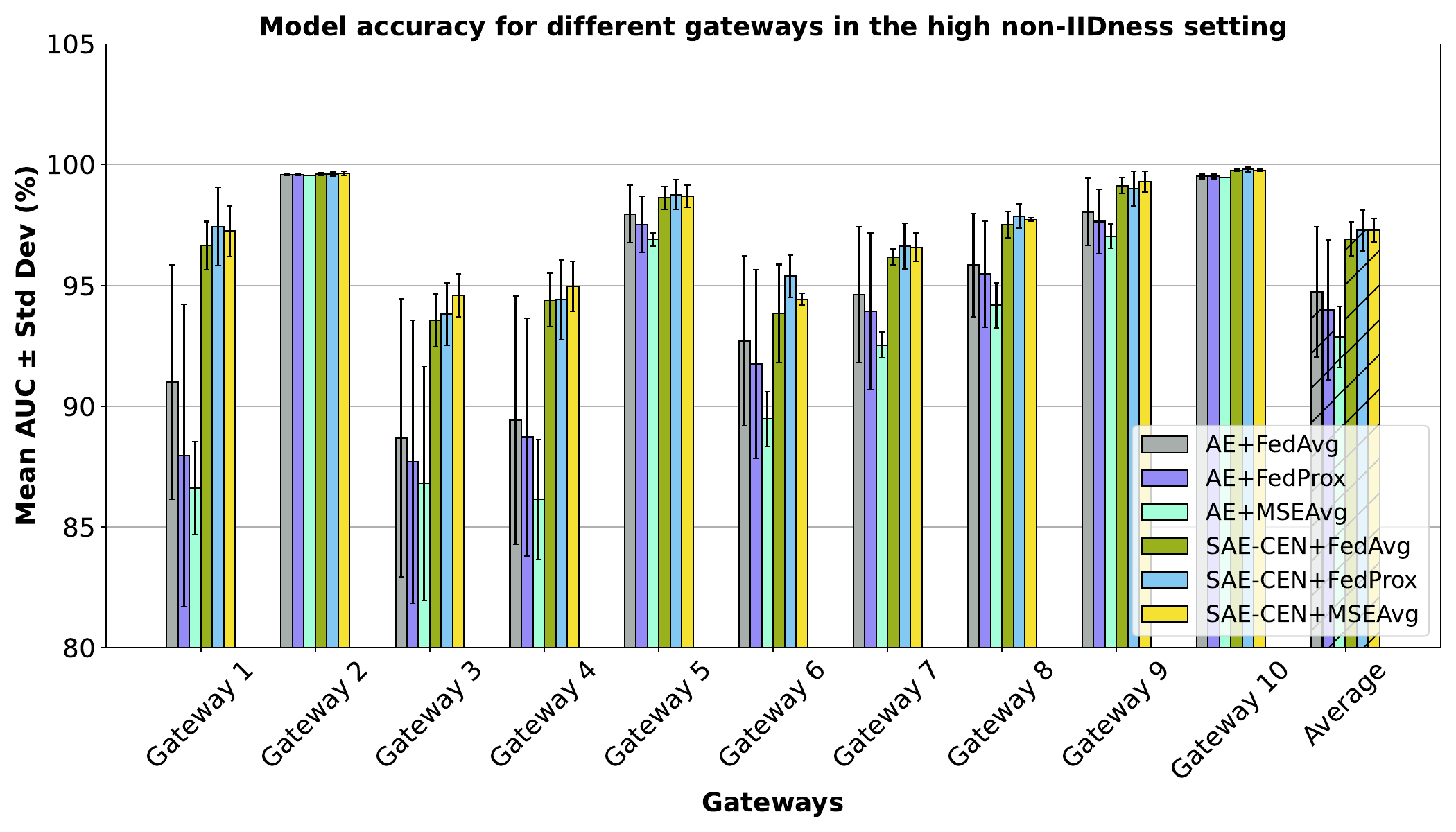}
  \captionsetup{justification=centering} 
  \caption{Accuracy for federated intrusion detection in the high non-IIDness setting}
  \label{fig:noniid-exp1}
\end{figure}

We do the first experiment to examine the detection ability of our approach in every network gateway. Table \ref{tab:exp1-result} and Figures \ref{fig:iid-exp1}, \ref{fig:noniid-exp1} present the AUCs obtained by the Autoencoder and SAE-CEN models under three aggregation algorithms. The results indicate that the larger non-IIDness setting poses greater challenges for machine learning models in detecting anomalies, leading to a decrease in accuracy and less consistency among gateways compared to the low non-IIDness scenario.

In the less heterogeneous case, both the Autoencoder and SAE-CEN models have high accuracy for all gateways under all federated learning algorithms, and the Autoencoder is slightly better than SAE-CEN in some algorithms. 
This suggests that the Autoencoder model can effectively generalize the whole data across gateways in this case. 
The reason may be due to the variability of training data among gateways is not much, all gateways have a similar data distribution, which helps Autoencoder to easily model the data. 
However, SAE-CEN does not outperform Autoencoder in all algorithms. 
The root cause is related to the architecture of the SAE model. During training, the SAE model must balance two objectives: representing the latent data close to the origin and reconstructing normal data. As a result, the SAE-CEN model cannot show outstanding performance compared to the Autoencoder model in the first scenario. This signifies that in a simple federated learning case, the Autoencoder model is good enough to detect anomalies accurately. 

\begin{figure}[!ht]
  \centering
  \includegraphics[width=\linewidth,height=0.40\textheight]{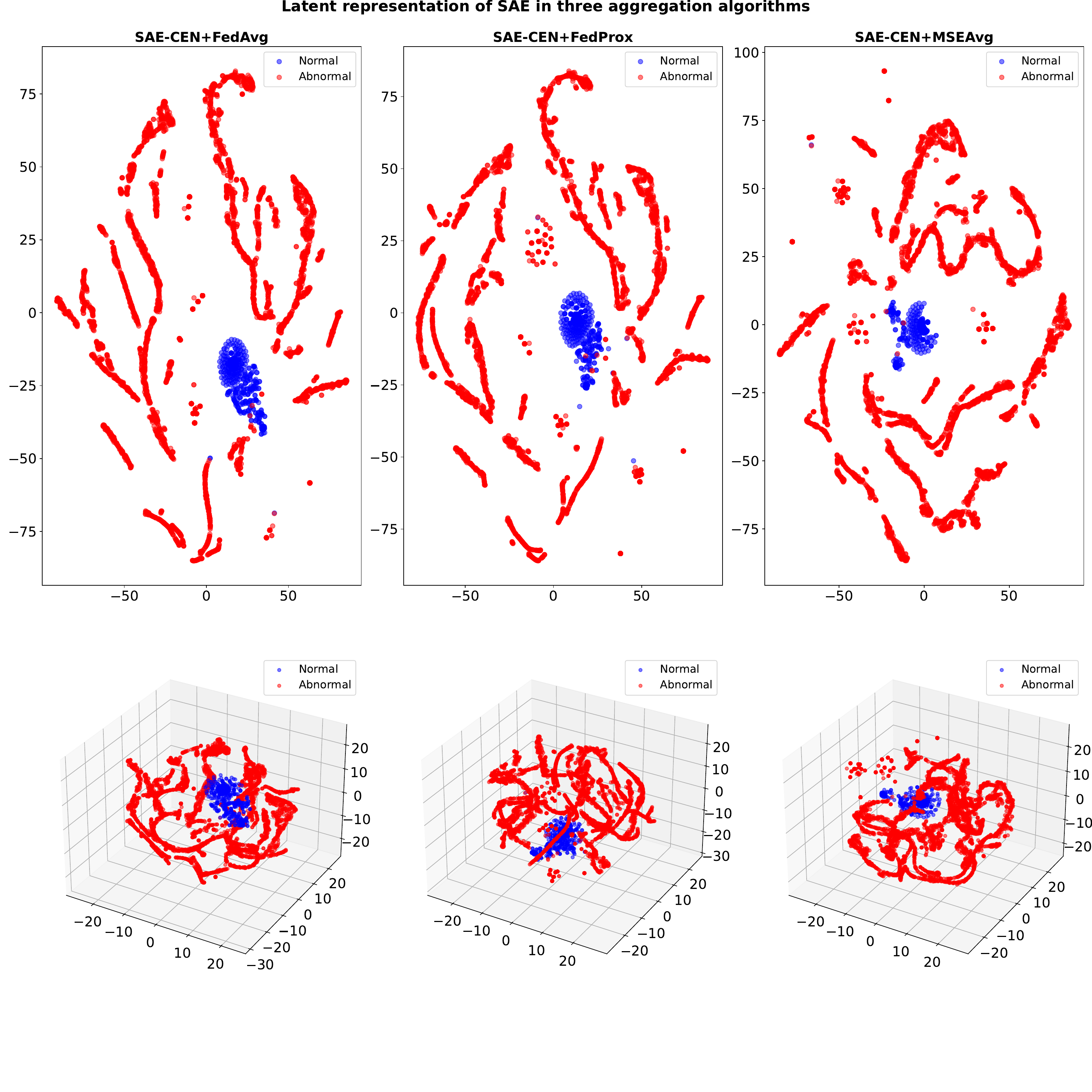} 
  \captionsetup{justification=centering} 
  \caption{Latent representation of SAE comparison in the federated learning schemes}
  \label{fig:latent}
\end{figure}

However, in the more heterogeneous setting, the Autoencoder accuracy drops dramatically under all aggregation algorithms. Meanwhile, SAE-CEN shows excellent effectiveness across all gateways. 
This can be explained as follows: The training data distribution across gateways is significantly different, and each local Autoencoder model converges in various directions. Thus, when accumulating all local models to a unique model and sending them to all gateways, the new model performance will drop substantially. 
In contrast, each local SAE-CEN model aims to optimize parameters to create a compact latent space for normal data when trying to force the normal latent data close to the origin. Therefore, the aggregated model will also represent the latent data approximately near the optimal area, enhancing the CEN model's ability to detect unseen data effectively.

The results highlight the improvements achieved by FedMSE, brought by the integration of the MSEAvg algorithm and the SAE-CEN machine learning model, particularly in the high non-IIDness case.
The smaller standard deviation also demonstrates improved consistency among gateways. FedAvg and FedProx update the global models based on the training data size of selected gateways. 
In the second setting, the gateway holding larger data may not ensure the representativeness of the whole data. For example, in this experiment setting (Figure \ref{fig:non-iid-dataset-training}), Gateway 5 may not be better than Gateway 6. 
Instead, MSEAvg enhances this by prioritizing updates from models that better model the representative of the whole data. By assigning greater weight to these accurate models, MSEAvg forces the aggregated weights close to the better models. This makes the global model improve the generalization ability and reduces the influence of noisy updates that might cause the abnormal data points to be mixed in the normal latent region. The aggregated model will act more effectively on all subnetworks.
Another reason is the behavior inside the SAE model, the SAE model needs to control the trade-off between reconstruction error and shrink error. MSEAvg helps the SAE model form the normal data comprehensively while keeping the latent representation capability. Therefore SAE model can separate the normal data points better than FedAvg and FedProx.

Figure \ref{fig:latent} visualizes the SAE's latent data on the testing dataset in 2D and 3D. The normal cluster that is constructed by MSEAvg is smaller, denser, and closer to the origin than that of FedAvg and FedProx algorithms. This helps the CEN model inside FedMSE work more effectively in detecting abnormal data points, particularly on unseen data when IoT networks innovate.

\subsection{Experiment 2: Effects of gateway selection ratio}
\label{exp:exp2}
In federated learning, the percentage of gateways chosen to participate in the training process can significantly affect the model's performance, efficiency, and communication overhead. There is always a trade-off among these criteria in choosing an optimal gateway selection ratio. We do the second experiment to consider FedMSE's efficiency in resource-constrained IoT networks.

\begin{figure}[ht]
  \centering
  \includegraphics[width=\columnwidth, height=0.19\textheight]{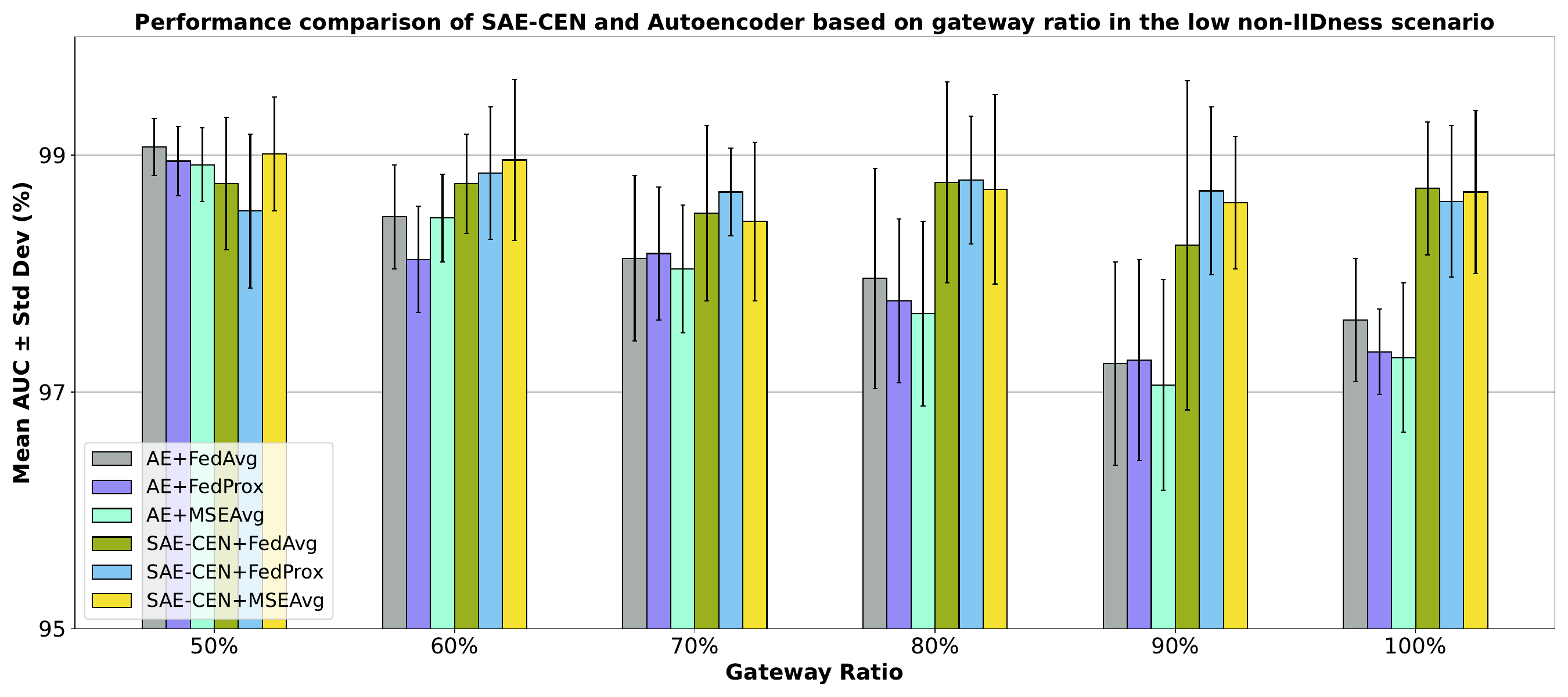}
  \captionsetup{justification=centering} 
  \caption{Accuracy for different gateway selection ratios in the low non-IIDness scenario}
  \label{fig:iid-exp2}
\end{figure}

\begin{figure}[ht]
  \centering
  \includegraphics[width=\columnwidth, height=0.19\textheight]{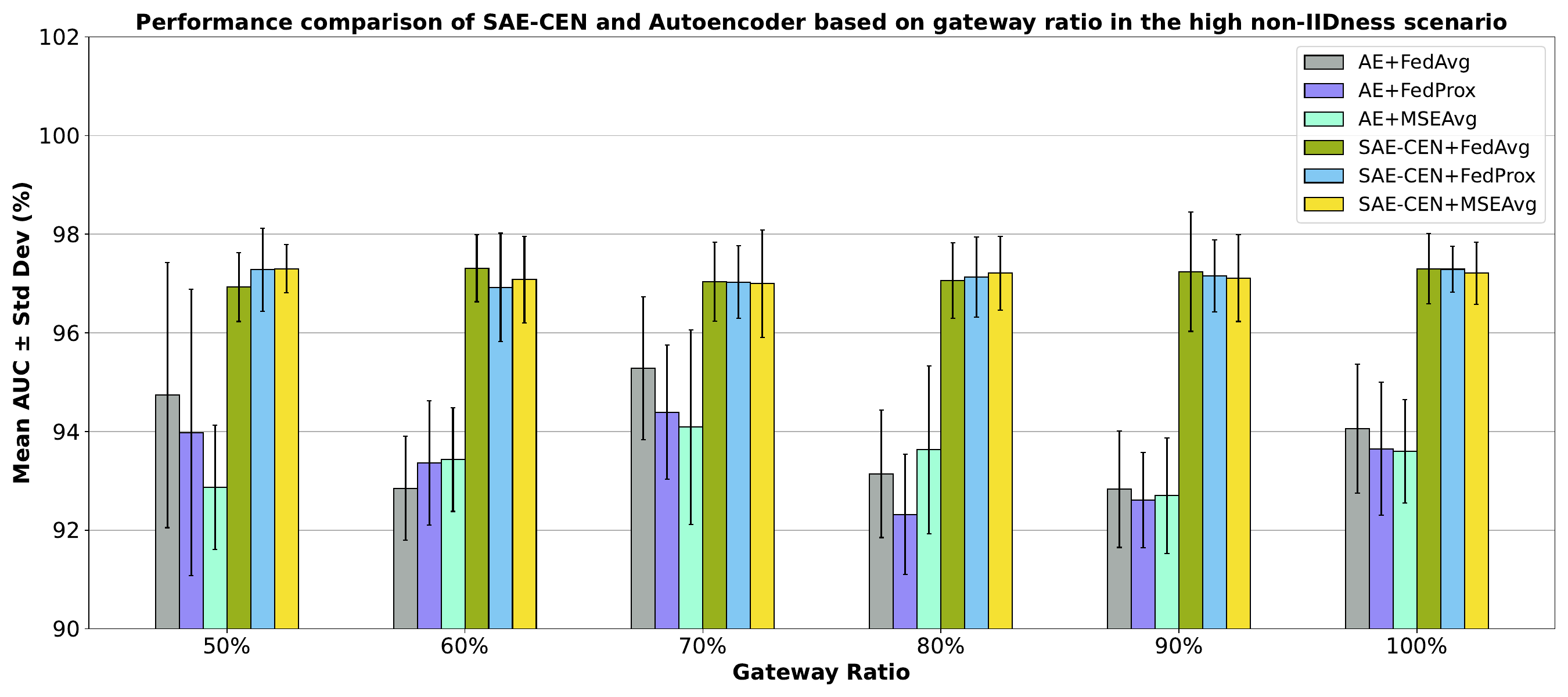}
  \captionsetup{justification=centering} 
  \caption{Accuracy for different gateway selection ratios in the high non-IIDness scenario}
  \label{fig:noniid-exp2}
\end{figure}

Table \ref{tab:combined-exp2} and Figure \ref{fig:iid-exp2}, \ref{fig:noniid-exp2} show the average accuracy of comparison candidates under different settings of the gateway selection ratio. The results indicate that the gateway selection ratio affects the consistency among gateways, and FedMSE still shows outstanding power. 
This is evidenced by changes in the standard deviation and the complexity of algorithms although the mean accuracy fluctuates slightly and all FedAvg, FedProx, and MSEAvg help SAE-CEN have comparable accuracy.

\begin{table*}[!ht]
\centering
\caption{Performance comparison (\%) based on gateway selection ratio in small and high non-IIDness scenarios. FedMSE results are represented by the combination of SAE-CEN with the MSEAvg algorithm.}
\label{tab:combined-exp2}
\resizebox{\textwidth}{!}{%
\begin{tabular}{@{}cccccccccc@{}}
\toprule
\multirow{2}{*}{\textbf{Gateway ratio}} & \multicolumn{3}{c}{\textbf{Autoencoder}} & \multicolumn{3}{c}{\textbf{SAE-CEN}} \\ \cmidrule(lr){2-4} \cmidrule(lr){5-7}
& \textbf{FedAvg} & \textbf{FedProx} & \textbf{MSEAvg} & \textbf{FedAvg} & \textbf{FedProx} & \textbf{MSEAvg} \\ \midrule
\multicolumn{7}{c}{\textit{a) Low non-IIDness}} \\ \midrule
\textbf{50\%}  & \textbf{99.07$\pm$0.24} & 98.95$\pm$0.29 & 98.92$\pm$0.31 & 98.76$\pm$0.56 & 98.53$\pm$0.65 & 99.01$\pm$0.48 \\
\textbf{60\%}  & 98.48$\pm$0.44 & 98.12$\pm$0.45 & 98.47$\pm$0.37 & 98.76$\pm$0.42 & 98.85$\pm$0.56 & \textbf{98.96$\pm$0.68} \\
\textbf{70\%}  & 98.13$\pm$0.70 & 98.17$\pm$0.56 & 98.04$\pm$0.54 & 98.51$\pm$0.74 & \textbf{98.69$\pm$0.37} & 98.44$\pm$0.67 \\
\textbf{80\%}  & 97.96$\pm$0.93 & 97.77$\pm$0.69 & 97.66$\pm$0.78 & 98.77$\pm$0.85 & \textbf{98.79$\pm$0.54} & 98.71$\pm$0.80 \\
\textbf{90\%}  & 97.24$\pm$0.86 & 97.27$\pm$0.85 & 97.06$\pm$0.89 & 98.24$\pm$1.39 & \textbf{98.70$\pm$0.71} & 98.60$\pm$0.56 \\
\textbf{100\%} & 97.61$\pm$0.52 & 97.34$\pm$0.36 & 97.29$\pm$0.63 & \textbf{98.72$\pm$0.56} & 98.61$\pm$0.64 & 98.69$\pm$0.69 \\  \midrule
\multicolumn{7}{c}{\textit{b) High non-IIDness}} \\ \midrule
\textbf{50\%}  & 94.74$\pm$2.69 & 93.98$\pm$2.90 & 92.87$\pm$1.26 & 96.93$\pm$0.70 & 97.28$\pm$0.84 & \textbf{97.30$\pm$0.49} \\
\textbf{60\%}  & 92.85$\pm$1.05 & 93.36$\pm$1.26 & 93.43$\pm$1.05 & \textbf{97.31$\pm$0.68} & 96.92$\pm$1.10 & 97.08$\pm$0.88 \\
\textbf{70\%}  & 95.28$\pm$1.45 & 94.39$\pm$1.36 & 94.09$\pm$1.97 & \textbf{97.04$\pm$0.80} & 97.03$\pm$0.74 & 97.00$\pm$1.09 \\
\textbf{80\%}  & 93.14$\pm$1.29 & 92.32$\pm$1.22 & 93.63$\pm$1.70 & 97.06$\pm$0.77 & 97.13$\pm$0.81 & \textbf{97.21$\pm$0.75} \\
\textbf{90\%}  & 92.83$\pm$1.18 & 92.61$\pm$0.97 & 92.70$\pm$1.17 & \textbf{97.24$\pm$1.21} & 97.15$\pm$0.73 & 97.11$\pm$0.88 \\
\textbf{100\%} & 94.06$\pm$1.31 & 93.65$\pm$1.35 & 93.60$\pm$1.05 & \textbf{97.30$\pm$0.71} & 97.29$\pm$0.46 & 97.21$\pm$0.63 \\ \bottomrule
\end{tabular}%
}
\end{table*}

Considering the low non-IIDness scenario, data is uniformly distributed among gateways. Therefore, each gateway has similar data characteristics, ensuring that even a subset of gateways can provide a representative sample for model training. Therefore, the standard deviation is relatively low for all gateway ratios, reflecting consistent performance among gateways. 
Autoencoder is sensitive to noise, hence, it tends to drop accuracy when updated by more gateways. This is due to a bit of difference in the data distribution among gateways, making more noise and variability in aggregation compared to a small ratio. 
The SAE-CEN model performs very accurately and consistently in all settings because of the representation ability as discussed in Experiment 1, even in different ratios.

It is more clear in the more heterogeneous case that the standard deviation is higher, indicating more variability in performance due to the non-uniform data distribution. The model accuracy has a trend similar to the low non-IIDness context, and the SAE-CEN model outperforms the Autoencoder model in all settings for both accuracy and consistency. The standard deviation of the Autoencoder model under the FedAvg and FedProx tends to be reduced when all gateways participate in the update process. This behavior occurs because including all gateways ensures that the global model is updated with the complete data distribution, enhancing generalization and leading to smaller standard deviations. The MSEAvg algorithm shows relatively lower standard deviations compared to FedAvg and FedProx, suggesting more consistent accuracy among gateways. Currently, we randomly select the gateways, which does not ensure the representativeness of the gateways in the population. Therefore, with smaller gateway selection ratios, there is a risk of overfitting due to the bias toward the gateways that are selected more times than others in the training process, which causes high variance across gateways in the IoT network. However, effective aggregation strategies like MSEAvg can partially offset this by updating the global model based on a representative dataset, the development dataset, and prioritizing more accurate local models.
This makes the global model act more consistently across gateways and ensures its generalization in small gateway ratios.

The computational costs aspect also needs to be rigorously considered in the IoT network, where IoT gateways do not have powerful resources. Higher gateway ratios mean more gateways are participating in each training round, increasing the computational load, communication overhead, and training time. Thus, it is critical to select an efficient model and gateway ratio. 
In the first scenario, choosing Autoencoder and FedAvg as a solution for IoT federated intrusion detection is good enough for even a small gateway ratio setting as 50\%.
However, in more practically heterogeneous scenarios, the SAE-CEN model is the best selection. 
FedAvg and FedProx also make SAE-CEN have high accuracy in some settings because they can leverage both the information of the data size and the data characteristics to update the global model with more participants. 
Nevertheless, FedMSE, which combines SAE-CEN and MSEAvg, can produce the highest accuracy and consistency even only with the smallest ratio, half of all gateways. This achieves the goal of both performance and computation expenses.
In both settings, it is enough to use a small ratio, with 50\% of all gateways participating in the training process with suitable models. This improves federated intrusion detection in all mentioned criteria.

Overall, we conclude that our approach, FedMSE, is the better solution for practical cases.

\subsection{Experiment 3: Accuracy for different IoT network sizes}
\label{exp:exp3}

\begin{table*}[ht]
\centering
\caption{Performance comparison (\%) based on network scales in small and high non-IIDness scenarios. FedMSE results are represented by the combination of SAE-CEN with the MSEAvg algorithm.}
\label{tab:combined-exp3}
\resizebox{\textwidth}{!}{%
\begin{tabular}{@{}cccccccccc@{}}
\toprule
\multirow{2}{*}{\textbf{Network scale}} & \multicolumn{3}{c}{\textbf{Autoencoder}} & \multicolumn{3}{c}{\textbf{SAE-CEN}} \\ \cmidrule(lr){2-4} \cmidrule(lr){5-7}
& \textbf{FedAvg} & \textbf{FedProx} & \textbf{MSEAvg} & \textbf{FedAvg} & \textbf{FedProx} & \textbf{MSEAvg} \\ \midrule
\multicolumn{7}{c}{\textit{a) Low non-IIDness}} \\ \midrule
\textbf{10-gateway} & \textbf{99.07$\pm$0.24} & 98.95$\pm$0.29 & 98.92$\pm$0.31 & 98.76$\pm$0.56 & 98.53$\pm$0.65 & 99.01$\pm$0.48 \\
\textbf{20-gateway} & 98.43$\pm$0.26 & 98.51$\pm$0.21 & 98.56$\pm$0.31 & 98.59$\pm$0.37 & \textbf{99.02$\pm$0.39} & 98.54$\pm$0.37 \\
\textbf{30-gateway} & 97.40$\pm$0.67 & 97.45$\pm$0.47 & 97.37$\pm$0.57 & 98.27$\pm$0.43 & \textbf{98.34$\pm$0.50} & 98.34$\pm$0.55 \\
\textbf{40-gateway} & 97.05$\pm$0.97 & 97.16$\pm$0.66 & 96.95$\pm$0.95 & 98.38$\pm$0.42 & 98.38$\pm$0.42 & \textbf{98.45$\pm$0.39} \\
\textbf{50-gateway} & 97.01$\pm$0.75 & 96.74$\pm$0.56 & 97.20$\pm$0.82 & \textbf{98.33$\pm$0.63} & 98.16$\pm$0.37 & 98.20$\pm$0.56 \\ \midrule
\multicolumn{7}{c}{\textit{b) High non-IIDness}} \\ \midrule
\textbf{10-gateway} & 94.74$\pm$2.69 & 93.98$\pm$2.90 & 92.87$\pm$1.26 & 96.93$\pm$0.70 & 97.28$\pm$0.84 & \textbf{97.30$\pm$0.49} \\
\textbf{20-gateway} & 95.72$\pm$1.42 & 95.62$\pm$1.18 & 95.68$\pm$1.11 & 97.17$\pm$0.90 & 97.19$\pm$0.83 & \textbf{97.29$\pm$0.96} \\
\textbf{30-gateway} & 96.95$\pm$1.51 & 96.96$\pm$1.19 & 96.98$\pm$1.43 & 97.54$\pm$0.61 & 97.65$\pm$0.54 & \textbf{97.73$\pm$0.56} \\
\textbf{40-gateway} & 95.99$\pm$1.83 & 95.78$\pm$1.78 & 95.72$\pm$1.80 & \textbf{97.81$\pm$0.94} & 97.58$\pm$0.84 & 97.77$\pm$1.01 \\
\textbf{50-gateway} & 96.30$\pm$0.98 & 96.39$\pm$0.85 & 96.42$\pm$2.37 & 98.41$\pm$0.72 & 98.36$\pm$0.65 & \textbf{98.52$\pm$0.51} \\ \bottomrule
\end{tabular}%
}
\end{table*}

\begin{figure}[!t]
  \centering
  \includegraphics[width=\columnwidth, height=0.19\textheight]{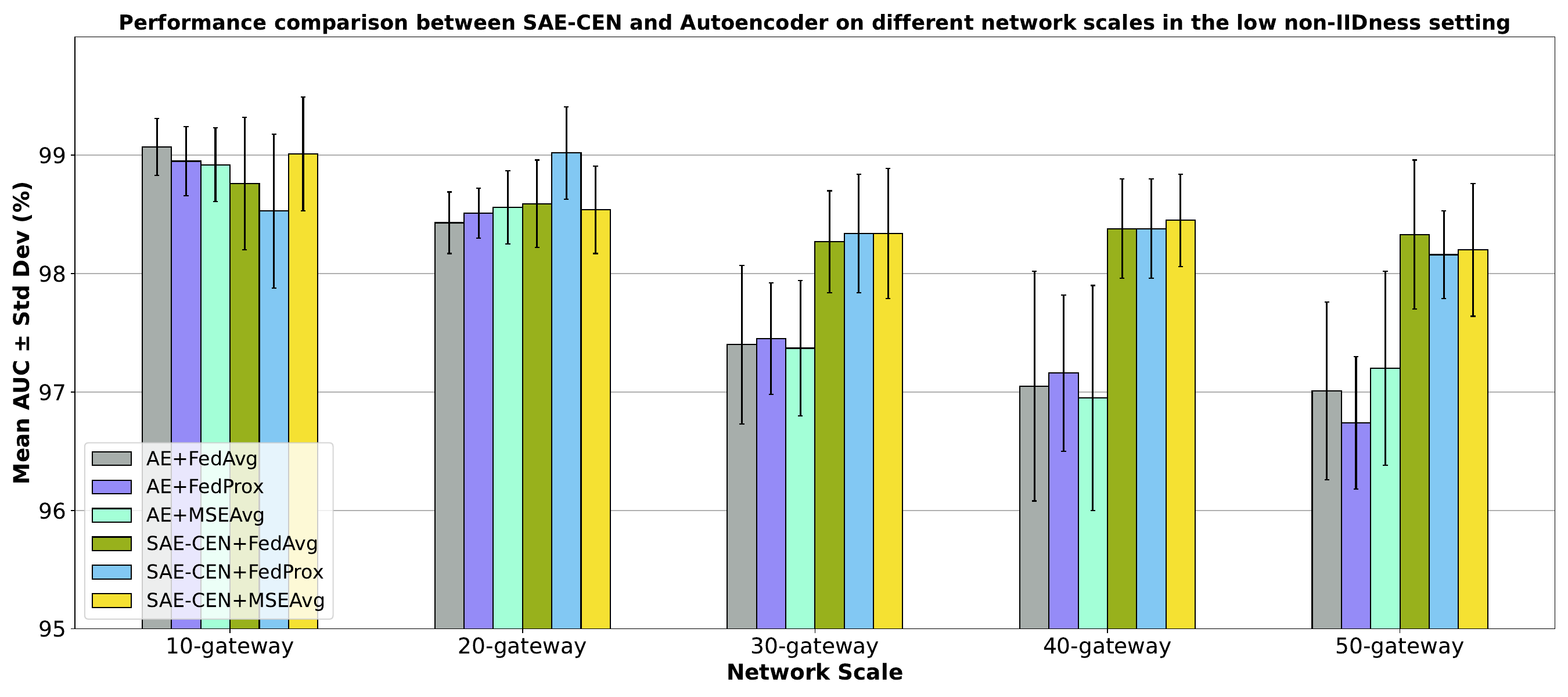}
  \captionsetup{justification=centering} 
  \caption{Accuracy for different IoT network sizes in the low non-IIDness scenario}
  \label{fig:iid-exp3}
\end{figure}

\begin{figure}[!t]
  \centering
  \includegraphics[width=\columnwidth, height=0.19\textheight]{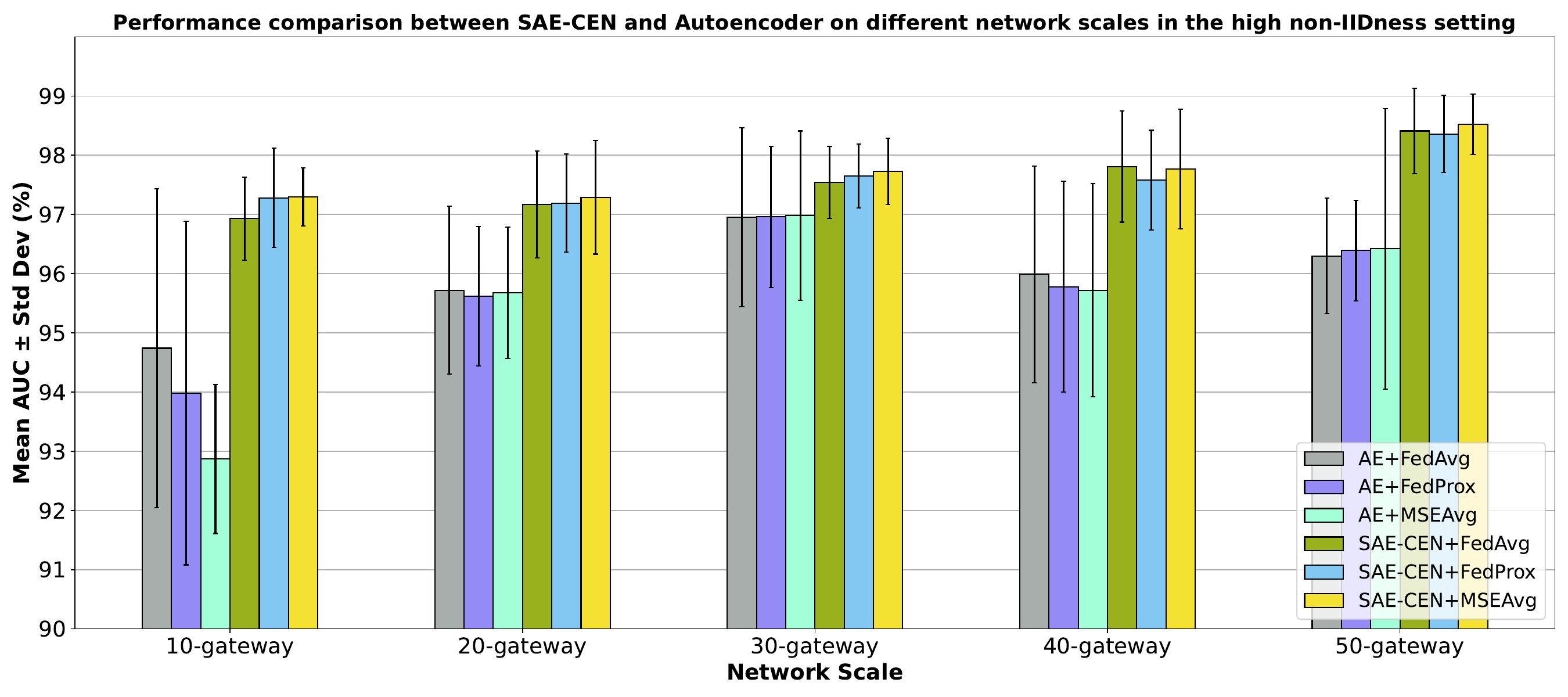}
  \captionsetup{justification=centering} 
  \caption{Accuracy for different IoT network sizes in the high non-IIDness scenario}
  \label{fig:noniid-exp3}
\end{figure}

Due to IoT networks changing in size extensively from the initial state during use time, it is necessary to choose a robust model for the dynamic characteristics of networks. This experiment is performed to investigate our approach's robustness in different network scales. Table \ref{tab:combined-exp3} and Figures~\ref{fig:iid-exp3} and \ref{fig:noniid-exp3} show the models' average performance under different settings of IoT network scale with the gateway ratio of 50\% and the similar hyperparameters to Experiment 1. 

When IoT networks scale, federated learning requires careful adjustment to balance the variability ratio among gateways and the ability to generalize knowledge effectively. In this research, we use an IoT device set that includes nine types of IoT appliances to illustrate practical IoT networks. The data partitions generated by the same IoT device, also known as the same distribution, may be similar. 
Thus, in the same gateway ratio setting and fixed device set, the selected data in a larger network provides a better representation of the entire network compared to a smaller network. However, the noise ratio in the aggregation process may be increased by more participants, leading to instability in federated learning.

In the low non-IIDness context, the Autoencoder performance tends to decrease with the expansion of the network.
The cause may be because of a bit of difference in data distribution controlled by Dirichlet distribution as discussed in Experiment 2 and the sensitivity of the Autoencoder model to noise. When the network grows, more gateways join in the training process, leading to a gradual increase in noise and variability. Meanwhile, it can not improve the generalization by the reason of the uniform distribution among gateways. This implies that Autoencoder can not perform effectively in large-scale networks, even in less heterogeneous settings.
In contrast, with the powerful representation capability, SAE-CEN still maintains outstanding performance.

The effectiveness of FedMSE is more clear in the high non-IIDness case when it shows the excellent and highest accuracy in most cases (Table \ref{tab:combined-exp3}). Furthermore, the accuracy of SAE-CEN demonstrates an upward trend as the network scale increases. Due to the heterogeneity of training data among gateways, more joining gateways bring more knowledge to the global model. This enables the SAE-CEN model to more effectively learn the normal data patterns across the entire network, especially under MSEAvg. Hence, the accuracy of SAE-CEN improves when the network size increases. 

The results of this experiment indicate that our proposed approach still possesses an outstanding power for large-scale networks, particularly in heterogeneous scenarios.

%% file: Sections/6-Conclusion.tex
\section{Conclusion}
\label{sec:conclusion}
This article presents a novel approach to enhance intrusion detection in IoT networks using federated learning, addressing the limitations of traditional centralized machine learning methods, such as privacy concerns, communication overheads, and data availability. 
We introduce a semi-supervised federated learning approach employing the Autoencoder-based model to improve detection accuracy and robustness in heterogeneous IoT networks. 
Major contributions include the proposal of a hybrid model combining the Shrink Autoencoder and the Centroid algorithm (SAE-CEN) in federated learning, and the development of a novel mean squared error-based aggregation algorithm (MSEAvg), which tackles challenges like data heterogeneity, unbalanced distributions, resource constraints, and the dynamic nature of IoT networks.

Experimental results using the N-BaIoT dataset demonstrate the effectiveness and efficiency of the proposed approach in real-world scenarios, achieving an AUC score of 97.30$\pm$0.49. 
This outperforms the combination of Autoencoder and FedAvg, which achieved 94.74$\pm$2.69, and Autoencoder with FedProx, which achieved 93.98$\pm$2.90. 
In terms of computational complexity, our approach significantly reduces resource consumption and communication overhead by involving only 50\% of gateways in the training process, while maintaining robustness in large-scale IoT networks.

However, the approach has several limitations, including sensitivity to hyperparameters and the need for more efficient gateway selection strategies in federated learning settings with heterogeneous data. The MSEAvg algorithm also increases computational costs due to additional server-side computations. The number of cyber threats is still limited with only Mirai and Gafgyt botnets. Future work should focus on implementing an experimental end-to-end framework to evaluate performance in practical networks, integrating techniques like hierarchy learning to enhance the aggregation algorithm, and optimizing hyperparameters for real-world applications. Furthermore, there should be additional experiments on various IoT network datasets. Addressing these challenges can further refine the proposed framework to better meet the evolving security needs of IoT networks.

%% file: main.bbl
\begin{thebibliography}{28}
\expandafter\ifx\csname natexlab\endcsname\relax\def\natexlab#1{#1}\fi
\providecommand{\url}[1]{\texttt{#1}}
\providecommand{\href}[2]{#2}
\providecommand{\path}[1]{#1}
\providecommand{\DOIprefix}{doi:}
\providecommand{\ArXivprefix}{arXiv:}
\providecommand{\URLprefix}{URL: }
\providecommand{\Pubmedprefix}{pmid:}
\providecommand{\doi}[1]{\href{http://dx.doi.org/#1}{\path{#1}}}
\providecommand{\Pubmed}[1]{\href{pmid:#1}{\path{#1}}}
\providecommand{\bibinfo}[2]{#2}
\ifx\xfnm\relax \def\xfnm[#1]{\unskip,\space#1}\fi
\bibitem[{Abdulrahman et~al.(2021)Abdulrahman, Tout, Ould-Slimane, Mourad, Talhi and Guizani}]{fl-survey}
\bibinfo{author}{Abdulrahman, S.}, \bibinfo{author}{Tout, H.}, \bibinfo{author}{Ould-Slimane, H.}, \bibinfo{author}{Mourad, A.}, \bibinfo{author}{Talhi, C.}, \bibinfo{author}{Guizani, M.}, \bibinfo{year}{2021}.
\newblock \bibinfo{title}{A survey on federated learning: The journey from centralized to distributed on-site learning and beyond}.
\newblock \bibinfo{journal}{IEEE Internet of Things Journal} \bibinfo{volume}{8}, \bibinfo{pages}{5476--5497}.
\newblock \DOIprefix\doi{10.1109/JIOT.2020.3030072}.
\bibitem[{Breunig et~al.(2000)Breunig, Kriegel, Ng and Sander}]{lof}
\bibinfo{author}{Breunig, M.M.}, \bibinfo{author}{Kriegel, H.P.}, \bibinfo{author}{Ng, R.T.}, \bibinfo{author}{Sander, J.}, \bibinfo{year}{2000}.
\newblock \bibinfo{title}{Lof: identifying density-based local outliers}, in: \bibinfo{booktitle}{Proceedings of the 2000 ACM SIGMOD International Conference on Management of Data}, \bibinfo{publisher}{Association for Computing Machinery}, \bibinfo{address}{New York, NY, USA}. p. \bibinfo{pages}{93–104}.
\newblock \URLprefix \url{https://doi.org/10.1145/342009.335388}, \DOIprefix\doi{10.1145/342009.335388}.
\bibitem[{Cao et~al.(2016)Cao, Nicolau and McDermott}]{ruleofthump}
\bibinfo{author}{Cao, V.L.}, \bibinfo{author}{Nicolau, M.}, \bibinfo{author}{McDermott, J.}, \bibinfo{year}{2016}.
\newblock \bibinfo{title}{A hybrid autoencoder and density estimation model for anomaly detection}, in: \bibinfo{editor}{Handl, J.}, \bibinfo{editor}{Hart, E.}, \bibinfo{editor}{Lewis, P.R.}, \bibinfo{editor}{L{\'o}pez-Ib{\'a}{\~{n}}ez, M.}, \bibinfo{editor}{Ochoa, G.}, \bibinfo{editor}{Paechter, B.} (Eds.), \bibinfo{booktitle}{Parallel Problem Solving from Nature -- PPSN XIV}, \bibinfo{publisher}{Springer International Publishing}, \bibinfo{address}{Cham}. pp. \bibinfo{pages}{717--726}.
\bibitem[{Cao et~al.(2019)Cao, Nicolau and McDermott}]{sae-model}
\bibinfo{author}{Cao, V.L.}, \bibinfo{author}{Nicolau, M.}, \bibinfo{author}{McDermott, J.}, \bibinfo{year}{2019}.
\newblock \bibinfo{title}{Learning neural representations for network anomaly detection}.
\newblock \bibinfo{journal}{IEEE Transactions on Cybernetics} \bibinfo{volume}{49}, \bibinfo{pages}{3074--3087}.
\newblock \DOIprefix\doi{10.1109/TCYB.2018.2838668}.
\bibitem[{Chandola et~al.(2009)Chandola, Banerjee and Kumar}]{anomaly-detection-survey}
\bibinfo{author}{Chandola, V.}, \bibinfo{author}{Banerjee, A.}, \bibinfo{author}{Kumar, V.}, \bibinfo{year}{2009}.
\newblock \bibinfo{title}{Anomaly detection: A survey}.
\newblock \bibinfo{journal}{ACM Comput. Surv.} \bibinfo{volume}{41}.
\newblock \URLprefix \url{https://doi.org/10.1145/1541880.1541882}, \DOIprefix\doi{10.1145/1541880.1541882}.
\bibitem[{Derawi et~al.(2020)Derawi, Dalveren and Cheikh}]{iot-transportation}
\bibinfo{author}{Derawi, M.}, \bibinfo{author}{Dalveren, Y.}, \bibinfo{author}{Cheikh, F.A.}, \bibinfo{year}{2020}.
\newblock \bibinfo{title}{Internet-of-things-based smart transportation systems for safer roads}, in: \bibinfo{booktitle}{2020 IEEE 6th World Forum on Internet of Things (WF-IoT)}, pp. \bibinfo{pages}{1--4}.
\newblock \DOIprefix\doi{10.1109/WF-IoT48130.2020.9221208}.
\bibitem[{Goodfellow et~al.(2016)Goodfellow, Bengio and Courville}]{deeplearningbook}
\bibinfo{author}{Goodfellow, I.}, \bibinfo{author}{Bengio, Y.}, \bibinfo{author}{Courville, A.}, \bibinfo{year}{2016}.
\newblock \bibinfo{title}{Deep Learning}.
\newblock \bibinfo{publisher}{MIT Press}.
\newblock \bibinfo{note}{\url{http://www.deeplearningbook.org}}.
\bibitem[{Guendouzi et~al.(2023)Guendouzi, Ouchani, {EL Assaad} and {EL Zaher}}]{fl-review}
\bibinfo{author}{Guendouzi, B.S.}, \bibinfo{author}{Ouchani, S.}, \bibinfo{author}{{EL Assaad}, H.}, \bibinfo{author}{{EL Zaher}, M.}, \bibinfo{year}{2023}.
\newblock \bibinfo{title}{A systematic review of federated learning: Challenges, aggregation methods, and development tools}.
\newblock \bibinfo{journal}{Journal of Network and Computer Applications} \bibinfo{volume}{220}, \bibinfo{pages}{103714}.
\newblock \URLprefix \url{https://www.sciencedirect.com/science/article/pii/S1084804523001339}, \DOIprefix\doi{https://doi.org/10.1016/j.jnca.2023.103714}.
\bibitem[{Gutierrez et~al.(2024)Gutierrez, Anagnostopoulos, Chatzigiannakis and Vitaletti}]{fedart-ml}
\bibinfo{author}{Gutierrez, D.M.J.}, \bibinfo{author}{Anagnostopoulos, A.}, \bibinfo{author}{Chatzigiannakis, I.}, \bibinfo{author}{Vitaletti, A.}, \bibinfo{year}{2024}.
\newblock \bibinfo{title}{Fedartml: A tool to facilitate the generation of non-iid datasets in a controlled way to support federated learning research}.
\newblock \bibinfo{journal}{IEEE Access} \bibinfo{volume}{12}, \bibinfo{pages}{81004--81016}.
\newblock \DOIprefix\doi{10.1109/ACCESS.2024.3410026}.
\bibitem[{Huang and Ling(2005)}]{AUC}
\bibinfo{author}{Huang, J.}, \bibinfo{author}{Ling, C.}, \bibinfo{year}{2005}.
\newblock \bibinfo{title}{Using auc and accuracy in evaluating learning algorithms}.
\newblock \bibinfo{journal}{IEEE Transactions on Knowledge and Data Engineering} \bibinfo{volume}{17}, \bibinfo{pages}{299--310}.
\newblock \DOIprefix\doi{10.1109/TKDE.2005.50}.
\bibitem[{Idrissi et~al.(2023)Idrissi, Alami, {El Mahdaouy}, {El Mekki}, Oualil, Yartaoui and Berrada}]{Fed-ANIDS}
\bibinfo{author}{Idrissi, M.J.}, \bibinfo{author}{Alami, H.}, \bibinfo{author}{{El Mahdaouy}, A.}, \bibinfo{author}{{El Mekki}, A.}, \bibinfo{author}{Oualil, S.}, \bibinfo{author}{Yartaoui, Z.}, \bibinfo{author}{Berrada, I.}, \bibinfo{year}{2023}.
\newblock \bibinfo{title}{Fed-anids: Federated learning for anomaly-based network intrusion detection systems}.
\newblock \bibinfo{journal}{Expert Systems with Applications} \bibinfo{volume}{234}, \bibinfo{pages}{121000}.
\newblock \URLprefix \url{https://www.sciencedirect.com/science/article/pii/S0957417423015026}, \DOIprefix\doi{https://doi.org/10.1016/j.eswa.2023.121000}.
\bibitem[{Kingma and Ba(2014)}]{adamoptimizer}
\bibinfo{author}{Kingma, D.P.}, \bibinfo{author}{Ba, J.}, \bibinfo{year}{2014}.
\newblock \bibinfo{title}{Adam: A method for stochastic optimization}.
\newblock \bibinfo{journal}{CoRR} \bibinfo{volume}{abs/1412.6980}.
\newblock \URLprefix \url{https://api.semanticscholar.org/CorpusID:6628106}.
\bibitem[{Li et~al.(2020)Li, Sahu, Zaheer, Sanjabi, Talwalkar and Smith}]{FedProx}
\bibinfo{author}{Li, T.}, \bibinfo{author}{Sahu, A.K.}, \bibinfo{author}{Zaheer, M.}, \bibinfo{author}{Sanjabi, M.}, \bibinfo{author}{Talwalkar, A.}, \bibinfo{author}{Smith, V.}, \bibinfo{year}{2020}.
\newblock \bibinfo{title}{Federated optimization in heterogeneous networks}, in: \bibinfo{booktitle}{Proceedings of Machine Learning and Systems}, pp. \bibinfo{pages}{429--450}.
\bibitem[{McMahan et~al.(2017)McMahan, Moore, Ramage, Hampson and Arcas}]{FL-proposal}
\bibinfo{author}{McMahan, B.}, \bibinfo{author}{Moore, E.}, \bibinfo{author}{Ramage, D.}, \bibinfo{author}{Hampson, S.}, \bibinfo{author}{Arcas, B.A.y.}, \bibinfo{year}{2017}.
\newblock \bibinfo{title}{{Communication-Efficient Learning of Deep Networks from Decentralized Data}}, in: \bibinfo{editor}{Singh, A.}, \bibinfo{editor}{Zhu, J.} (Eds.), \bibinfo{booktitle}{Proceedings of the 20th International Conference on Artificial Intelligence and Statistics}, \bibinfo{publisher}{PMLR}. pp. \bibinfo{pages}{1273--1282}.
\newblock \URLprefix \url{https://proceedings.mlr.press/v54/mcmahan17a.html}.
\bibitem[{Meidan et~al.(2018)Meidan, Bohadana, Mathov, Mirsky, Shabtai, Breitenbacher and Elovici}]{nbaiotdataset}
\bibinfo{author}{Meidan, Y.}, \bibinfo{author}{Bohadana, M.}, \bibinfo{author}{Mathov, Y.}, \bibinfo{author}{Mirsky, Y.}, \bibinfo{author}{Shabtai, A.}, \bibinfo{author}{Breitenbacher, D.}, \bibinfo{author}{Elovici, Y.}, \bibinfo{year}{2018}.
\newblock \bibinfo{title}{N-baiot—network-based detection of iot botnet attacks using deep autoencoders}.
\newblock \bibinfo{journal}{IEEE Pervasive Computing} \bibinfo{volume}{17}, \bibinfo{pages}{12--22}.
\newblock \DOIprefix\doi{10.1109/MPRV.2018.03367731}.
\bibitem[{Mothukuri et~al.(2022)Mothukuri, Khare, Parizi, Pouriyeh, Dehghantanha and Srivastava}]{gru-ensemble}
\bibinfo{author}{Mothukuri, V.}, \bibinfo{author}{Khare, P.}, \bibinfo{author}{Parizi, R.M.}, \bibinfo{author}{Pouriyeh, S.}, \bibinfo{author}{Dehghantanha, A.}, \bibinfo{author}{Srivastava, G.}, \bibinfo{year}{2022}.
\newblock \bibinfo{title}{Federated-learning-based anomaly detection for iot security attacks}.
\newblock \bibinfo{journal}{IEEE Internet of Things Journal} \bibinfo{volume}{9}, \bibinfo{pages}{2545--2554}.
\newblock \DOIprefix\doi{10.1109/JIOT.2021.3077803}.
\bibitem[{Nguyen et~al.(2019)Nguyen, Marchal, Miettinen, Fereidooni, Asokan and Sadeghi}]{diot}
\bibinfo{author}{Nguyen, T.}, \bibinfo{author}{Marchal, S.}, \bibinfo{author}{Miettinen, M.}, \bibinfo{author}{Fereidooni, H.}, \bibinfo{author}{Asokan, N.}, \bibinfo{author}{Sadeghi, A.}, \bibinfo{year}{2019}.
\newblock \bibinfo{title}{D{\"i}ot: A federated self-learning anomaly detection system for iot}, in: \bibinfo{booktitle}{2019 IEEE 39th International Conference on Distributed Computing Systems (ICDCS)}, \bibinfo{publisher}{IEEE}, \bibinfo{address}{United States}. pp. \bibinfo{pages}{756--767}.
\newblock \DOIprefix\doi{10.1109/ICDCS.2019.00080}. \bibinfo{note}{international Conference on Distributed Computing Systems , ICDCS ; Conference date: 07-07-2019 Through 10-07-2019}.
\bibitem[{Nguyen et~al.(2024)Nguyen, Ngo, Nguyen, Nguyen and Shone}]{VQuanNguyen-DNACE}
\bibinfo{author}{Nguyen, V.Q.}, \bibinfo{author}{Ngo, L.T.}, \bibinfo{author}{Nguyen, L.M.}, \bibinfo{author}{Nguyen, V.H.}, \bibinfo{author}{Shone, N.}, \bibinfo{year}{2024}.
\newblock \bibinfo{title}{Deep clustering hierarchical latent representation for anomaly-based cyber-attack detection}.
\newblock \bibinfo{journal}{Knowledge-Based Systems} \bibinfo{volume}{301}, \bibinfo{pages}{112366}.
\newblock \URLprefix \url{https://www.sciencedirect.com/science/article/pii/S0950705124010001}, \DOIprefix\doi{https://doi.org/10.1016/j.knosys.2024.112366}.
\bibitem[{Prechelt(2012)}]{earlystopping}
\bibinfo{author}{Prechelt, L.}, \bibinfo{year}{2012}.
\newblock \bibinfo{title}{Early Stopping --- But When?}. \bibinfo{publisher}{Springer Berlin Heidelberg}, \bibinfo{address}{Berlin, Heidelberg}.
\newblock pp. \bibinfo{pages}{53--67}.
\newblock \URLprefix \url{https://doi.org/10.1007/978-3-642-35289-8_5}, \DOIprefix\doi{10.1007/978-3-642-35289-8_5}.
\bibitem[{Rejeb et~al.(2023)Rejeb, Rejeb, Treiblmaier, Appolloni, Alghamdi, Alhasawi and Iranmanesh}]{iot-healcare}
\bibinfo{author}{Rejeb, A.}, \bibinfo{author}{Rejeb, K.}, \bibinfo{author}{Treiblmaier, H.}, \bibinfo{author}{Appolloni, A.}, \bibinfo{author}{Alghamdi, S.}, \bibinfo{author}{Alhasawi, Y.}, \bibinfo{author}{Iranmanesh, M.}, \bibinfo{year}{2023}.
\newblock \bibinfo{title}{The internet of things (iot) in healthcare: Taking stock and moving forward}.
\newblock \bibinfo{journal}{Internet of Things} \bibinfo{volume}{22}, \bibinfo{pages}{100721}.
\newblock \DOIprefix\doi{https://doi.org/10.1016/j.iot.2023.100721}.
\bibitem[{Rey et~al.(2022)Rey, {Sánchez Sánchez}, {Huertas Celdrán} and Bovet}]{FL-malware-iot}
\bibinfo{author}{Rey, V.}, \bibinfo{author}{{Sánchez Sánchez}, P.M.}, \bibinfo{author}{{Huertas Celdrán}, A.}, \bibinfo{author}{Bovet, G.}, \bibinfo{year}{2022}.
\newblock \bibinfo{title}{Federated learning for malware detection in iot devices}.
\newblock \bibinfo{journal}{Computer Networks} \bibinfo{volume}{204}, \bibinfo{pages}{108693}.
\newblock \URLprefix \url{https://www.sciencedirect.com/science/article/pii/S1389128621005582}, \DOIprefix\doi{https://doi.org/10.1016/j.comnet.2021.108693}.
\bibitem[{Sattler et~al.(2020)Sattler, Wiedemann, Müller and Samek}]{noniid-issue}
\bibinfo{author}{Sattler, F.}, \bibinfo{author}{Wiedemann, S.}, \bibinfo{author}{Müller, K.R.}, \bibinfo{author}{Samek, W.}, \bibinfo{year}{2020}.
\newblock \bibinfo{title}{Robust and communication-efficient federated learning from non-i.i.d. data}.
\newblock \bibinfo{journal}{IEEE Transactions on Neural Networks and Learning Systems} \bibinfo{volume}{31}, \bibinfo{pages}{3400--3413}.
\newblock \DOIprefix\doi{10.1109/TNNLS.2019.2944481}.
\bibitem[{Sharma et~al.(2023)Sharma, Sharma, Lal and Roy}]{iot-definition}
\bibinfo{author}{Sharma, B.}, \bibinfo{author}{Sharma, L.}, \bibinfo{author}{Lal, C.}, \bibinfo{author}{Roy, S.}, \bibinfo{year}{2023}.
\newblock \bibinfo{title}{Anomaly based network intrusion detection for iot attacks using deep learning technique}.
\newblock \bibinfo{journal}{Computers and Electrical Engineering} \bibinfo{volume}{107}, \bibinfo{pages}{108626}.
\newblock \URLprefix \url{https://www.sciencedirect.com/science/article/pii/S0045790623000514}, \DOIprefix\doi{https://doi.org/10.1016/j.compeleceng.2023.108626}.
\bibitem[{Vu et~al.(2022)Vu, Cao, Nguyen, Nguyen, Hoang and Dutkiewicz}]{iot-attack}
\bibinfo{author}{Vu, L.}, \bibinfo{author}{Cao, V.L.}, \bibinfo{author}{Nguyen, Q.U.}, \bibinfo{author}{Nguyen, D.N.}, \bibinfo{author}{Hoang, D.T.}, \bibinfo{author}{Dutkiewicz, E.}, \bibinfo{year}{2022}.
\newblock \bibinfo{title}{Learning latent representation for iot anomaly detection}.
\newblock \bibinfo{journal}{IEEE Transactions on Cybernetics} \bibinfo{volume}{52}, \bibinfo{pages}{3769--3782}.
\newblock \DOIprefix\doi{10.1109/TCYB.2020.3013416}.
\bibitem[{Wang et~al.(2023)Wang, Wang, Javaheri, Almutairi, Moghadamnejad and Younes}]{fl-iot}
\bibinfo{author}{Wang, X.}, \bibinfo{author}{Wang, Y.}, \bibinfo{author}{Javaheri, Z.}, \bibinfo{author}{Almutairi, L.}, \bibinfo{author}{Moghadamnejad, N.}, \bibinfo{author}{Younes, O.S.}, \bibinfo{year}{2023}.
\newblock \bibinfo{title}{Federated deep learning for anomaly detection in the internet of things}.
\newblock \bibinfo{journal}{Computers and Electrical Engineering} \bibinfo{volume}{108}, \bibinfo{pages}{108651}.
\newblock \URLprefix \url{https://www.sciencedirect.com/science/article/pii/S0045790623000769}, \DOIprefix\doi{https://doi.org/10.1016/j.compeleceng.2023.108651}.
\bibitem[{Yadav and Vishwakarma(2018)}]{iot-smartcity}
\bibinfo{author}{Yadav, P.}, \bibinfo{author}{Vishwakarma, S.}, \bibinfo{year}{2018}.
\newblock \bibinfo{title}{Application of internet of things and big data towards a smart city}, in: \bibinfo{booktitle}{2018 3rd International Conference On Internet of Things: Smart Innovation and Usages (IoT-SIU)}, pp. \bibinfo{pages}{1--5}.
\newblock \DOIprefix\doi{10.1109/IoT-SIU.2018.8519920}.
\bibitem[{Zhang et~al.(2015)Zhang, Xu and Gong}]{ocsvm}
\bibinfo{author}{Zhang, M.}, \bibinfo{author}{Xu, B.}, \bibinfo{author}{Gong, J.}, \bibinfo{year}{2015}.
\newblock \bibinfo{title}{An anomaly detection model based on one-class svm to detect network intrusions}, in: \bibinfo{booktitle}{2015 11th International Conference on Mobile Ad-hoc and Sensor Networks (MSN)}, pp. \bibinfo{pages}{102--107}.
\newblock \DOIprefix\doi{10.1109/MSN.2015.40}.
\bibitem[{Zhang et~al.(2023)Zhang, Mavromatis, Vafeas, Nejabati and Simeonidou}]{fed-feature-selection}
\bibinfo{author}{Zhang, X.}, \bibinfo{author}{Mavromatis, A.}, \bibinfo{author}{Vafeas, A.}, \bibinfo{author}{Nejabati, R.}, \bibinfo{author}{Simeonidou, D.}, \bibinfo{year}{2023}.
\newblock \bibinfo{title}{Federated feature selection for horizontal federated learning in iot networks}.
\newblock \bibinfo{journal}{IEEE Internet of Things Journal} \bibinfo{volume}{10}, \bibinfo{pages}{10095--10112}.
\newblock \DOIprefix\doi{10.1109/JIOT.2023.3237032}.

\end{thebibliography}
